\documentclass[sn-mathphys,Numbered]{sn-jnl}

\usepackage{array}
\usepackage{graphics}
\usepackage{graphicx}%
\usepackage{amsmath,amssymb,amsfonts}%
\usepackage{amsthm}%
\usepackage{mathrsfs}%
\usepackage[title]{appendix}%
\usepackage{xcolor}%
\usepackage{textcomp}%
\usepackage{manyfoot}%
\usepackage{booktabs}%
\usepackage{algorithm}%
\usepackage{algorithmicx}%
\usepackage{algpseudocode}%
\usepackage{listings}%
\usepackage{subfig}
\usepackage{colortbl}
\usepackage{bm}

\theoremstyle{plain}
\newtheorem{theorem}{Theorem}[section]

\theoremstyle{definition}
\newtheorem{definition}[theorem]{Definition}

\newtheorem{hypothesis}[theorem]{Hypothesis}
\theoremstyle{remark}

\begin{document}

\title[Geometry-Complete Diffusion for 3D Molecule Generation and Optimization]{Geometry-Complete Diffusion for 3D Molecule Generation and Optimization}


\author*[1]{\fnm{Alex} \sur{Morehead}}\email{acmwhb@missouri.edu}

\author[1]{\fnm{Jianlin} \sur{Cheng}}\email{chengji@missouri.edu}

\affil[1]{\orgdiv{Electrical Engineering \& Computer Science, NextGen Precision Health}, \orgname{University of Missouri}, \orgaddress{
\city{Columbia}, \postcode{65211}, \state{Missouri}, \country{USA}}}


\abstract{
\textbf{Motivation:} Generative deep learning methods have recently been proposed for generating 3D molecules using equivariant graph neural networks (GNNs) within a denoising diffusion framework. However, such methods are unable to learn important geometric properties of 3D molecules, as they adopt molecule-agnostic and non-geometric GNNs as their 3D graph denoising networks, which notably hinders their ability to generate valid large 3D molecules.

\textbf{Results:} In this work, we address these gaps by introducing the Geometry-Complete Diffusion Model (\textsc{GCDM}) for 3D molecule generation, which outperforms existing 3D molecular diffusion models by significant margins across conditional and unconditional settings for the QM9 dataset and the larger GEOM-Drugs dataset, respectively. Importantly, we demonstrate that \textsc{GCDM}'s generative denoising process enables the model to generate a significant proportion of valid and energetically-stable large molecules at the scale of GEOM-Drugs, whereas previous methods fail to do so with the features they learn. Additionally, we show that extensions of \textsc{GCDM} can not only effectively design 3D molecules for specific protein pockets but can be repurposed to consistently optimize the geometry and chemical composition of existing 3D molecules for molecular stability and property specificity, demonstrating new versatility of molecular diffusion models.

\textbf{Availability:} Code and data are freely available on \href{https://github.com/BioinfoMachineLearning/Bio-Diffusion}{GitHub}.
}

\keywords{Geometric deep learning, Diffusion generative modeling, 3D molecules}



\maketitle

\section{Introduction}\label{sec1}

Generative modeling has recently been experiencing a renaissance in modeling efforts driven largely by denoising diffusion probabilistic models (DDPMs). At a high level, DDPMs are trained by learning how to denoise a noisy version of an input example. For example, in the context of computer vision, Gaussian noise may be successively added to an input image with the goals of a DDPM in mind. We would then desire for a generative model of images to learn how to successfully distinguish between the original input image's feature signal and the noise added to the image thereafter. If a model can achieve such outcomes, we can use the model to generate novel images by first sampling multivariate Gaussian noise and then iteratively removing, from the current state of the image, the noise predicted by the model. This classic formulation of DDPMs has achieved significant results in the space of image generation \citep{rombach2022high}, audio synthesis \citep{kong2020diffwave}, and even meta-learning by learning how to conditionally generate neural network checkpoints \citep{peebles2022learning}. Furthermore, such an approach to generative modeling has expanded its reach to encompass scientific disciplines such as computational biology \citep{anand2022protein,corso2022diffdock,guo2023,watson2023novo,morehead2023towards}, computational chemistry \citep{xu2022geodiff, gebauer2022inverse, anstine2023generative}, and computational physics \citep{mudur2022can}.

\begin{figure}[t]
\centering
\includegraphics[width=\textwidth]{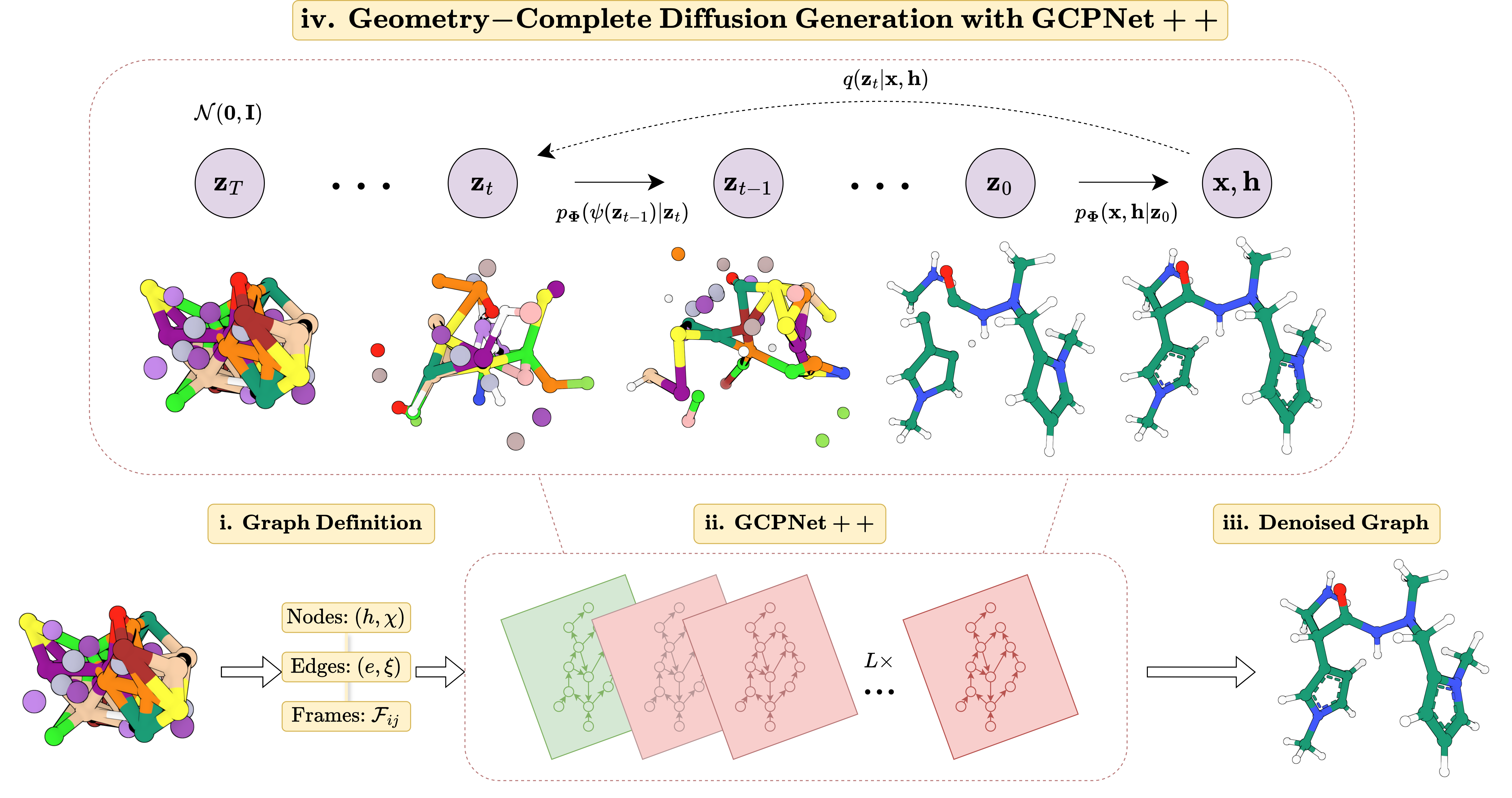}
\caption{A framework overview of the proposed \textit{Geometry-Complete Diffusion Model} (\textsc{GCDM}) for geometric and chirality-aware 3D molecule generation. The framework consists of (\textbf{i.}) a graph (topology) definition process; (\textbf{ii.}) a \textsc{GCPNet}-based graph neural network for \textrm{SE(3)}-equivariant graph representation learning; (\textbf{iii.}) denoising of 3D input graphs using \textsc{GCPNet++}; and (\textbf{iv.}) application of a trained \textsc{GCPNet++} denoising network for 3D molecule generation. Zoom in for the best viewing experience.}
\label{figure:gcdm}
\end{figure}

Concurrently, the field of geometric deep learning (GDL) \citep{bronstein2021geometric} has seen a sizeable increase in research interest lately, driven largely by theoretical advances within the discipline \citep{joshi2023expressive} as well as by novel applications of such methodology \citep{stark2022equibind, morehead2022geometric, jamasb2024evaluating, morehead2024gcpnet_ema}. Notably, such applications even include what is considered by many researchers to be a solution to the problem of predicting 3D protein structures from their corresponding amino acid sequences \citep{jumper2021highly}. Such an outcome arose, in part, from recent advances in sequence-based language modeling efforts \citep{vaswani2017attention, lin2023evolutionary} as well as from innovations in equivariant neural network modeling \citep{thomas2018tensor}.

However, it is currently unclear how the expressiveness of geometric neural networks impacts the ability of generative methods that incorporate them to faithfully model a geometric data distribution. In addition, it is currently unknown whether diffusion models for 3D molecules can be repurposed for important, real-world tasks without retraining or fine-tuning and whether geometric diffusion models are better equipped for such tasks. Toward this end, in this work, we provide the following findings.
\begin{itemize}
    \item Neural networks that perform message-passing with geometric quantities enable diffusion generative models of 3D molecules to generate valid and energetically-stable large molecules whereas non-geometric message-passing networks \underline{fail} to do so, where we introduce key computational metrics to enable such findings.
    \item Physical inductive biases such as invariant graph attention and molecular chirality \underline{both} play important roles in diffusion-generating valid 3D molecules.
    \item Our newly-proposed Geometry-Complete Diffusion Model (\textsc{GCDM}), which is the first diffusion model to incorporate the above insights and achieve the ideal type of equivariance for 3D molecule generation (i.e., \textrm{SE(3)} equivariance), establishes new state-of-the-art (SOTA) results for conditional 3D molecule generation on the QM9 dataset as well as for unconditional molecule generation on the GEOM-Drugs dataset of \underline{large} 3D molecules, for the latter more than doubling PoseBusters validity rates; generates more unique and novel small molecules for unconditional generation on the QM9 dataset; and achieves better Vina energy scores and more than twofold higher PoseBusters validity rates \citep{buttenschoen2024posebusters} for protein-conditioned 3D molecule generation.
    \item We further demonstrate that geometric diffusion models such as \textsc{GCDM} can consistently perform 3D molecule optimization for molecular stability as well as for specific molecular properties \underline{without} requiring any retraining and can consistently do so whereas non-geometric diffusion models cannot.
\end{itemize}

\section{Results}
\label{section:results}

\subsection{Unconditional 3D Molecule Generation - QM9}
\label{section:unconditional_qm9}
The first dataset used in our experiments, the QM9 dataset \citep{ramakrishnan2014quantum}, contains molecular properties and 3D atom coordinates for 130k small molecules. Each molecule in QM9 can contain up to 29 atoms after hydrogen atoms are imputed for each molecule following dataset postprocessing as in \citet{hoogeboom2022equivariant}. For the task of 3D molecule generation, we train \textsc{GCDM} to unconditionally generate molecules by producing atom types (H, C, N, O, and F), integer atom charges, and 3D coordinates for each of the molecules' atoms. Following \citet{anderson2019cormorant}, we split QM9 into training, validation, and test partitions consisting of 100k, 18k, and 13k molecule examples, respectively.

\textbf{Metrics.} We measure each method's average negative log-likelihood (NLL) over the corresponding test dataset, for methods that report this quantity. Intuitively, a method achieving a lower test NLL compared to other methods indicates that the method can more accurately predict \textit{denoised} pairings of atom types and coordinates for unseen data, implying that it has fit the underlying data distribution more precisely than other methods. In terms of molecule-specific metrics, we adopt the scoring conventions of \citet{satorras2021f} by using the distance between atom pairs and their respective atom types to predict bond types (single, double, triple, or none) for all but one baseline method (i.e., E-NF). Subsequently, we measure the proportion of generated atoms that have the right valency (atom stability - AS) and the proportion of generated molecules for which all atoms are stable (molecule stability - MS). To offer additional insights into each method's behavior for 3D molecule generation, we also report the validity (Val) of the generated molecules as determined by RDKit \citep{landrum2013rdkit}, the uniqueness of the generated molecules overall (Uniq), and whether the generated molecules pass each of the \textit{de novo} chemical and structural validity tests (i.e., sanitizable, all atoms connected, valid bond lengths and angles, no internal steric clashes, flat aromatic rings and double bonds, low internal energy, correct valence, and kekulizable) proposed in the PoseBusters software suite \citep{buttenschoen2024posebusters} and adopted by recent works on molecule generation tasks \citep{krishna2023generalized, deepmind2023alphafold}. Each method's results in the top half (bottom half) of Table \ref{table:3dmg_unconditional_qm9_results} are reported as the mean and standard deviation (mean and Student's t-distribution 95\% confidence error intervals) ($\pm$) of each metric across three (five) test runs on QM9, respectively.

\textbf{Baselines.} Besides including a reference point for molecule quality metrics using QM9 itself (i.e., Data), we compare \textsc{GCDM} (a geometry-complete DDPM - i.e., GC-DDPM) to 10 baseline models for 3D molecule generation, each trained and tested using the same corresponding QM9 splits for fair comparisons: G-Schnet \citep{gebauer2019symmetry}; Equivariant Normalizing Flows (E-NF) \citep{satorras2021f}; Graph Diffusion Models (GDM) \citep{hoogeboom2022equivariant} and their variations (i.e., GCM-aug); Equivariant Diffusion Models (EDM) \citep{hoogeboom2022equivariant}; Bridge and Bridge + Force \citep{wu2022diffusion}; latent diffusion models (LDMs) such as GraphLDM and its variation GraphLDM-aug \citep{xu2023geometric}; as well as the state-of-the-art GeoLDM method \citep{xu2023geometric}. Note that we specifically include these baselines as representative \textit{implicit bond prediction} methods for which bonds are inferred using their generated molecules' atom types and inter-atom distances, in contrast to \textit{explicit bond prediction} approaches such as those of \cite{vignac2023midi} and \cite{le2023navigating} for fair comparisons with our method. For each of such baseline methods, we report their results as curated by \citet{wu2022diffusion} and \citet{xu2023geometric}. We further include two \textsc{GCDM} ablation models to more closely analyze the impact of certain key model components within \textsc{GCDM}. These two ablation models include \textsc{GCDM} without chiral and geometry-complete local frames $\mathcal{F}_{ij}$ (i.e., \textsc{GCDM} w/o Frames) and \textsc{GCDM} without scalar message attention (SMA) applied to each edge message (i.e., \textsc{GCDM} w/o SMA). In Section \ref{section:methods} as well as Appendices \ref{section:appendix_diffusion_models} and \ref{section:appendix_additional_details}, we further discuss \textsc{GCDM}'s design, hyperparameters, and optimization with these model configurations.

\begin{table}[t]
\centering
\resizebox{\textwidth}{!}{%
    \begin{tabular}{lllllll}
        & & & & \\ \midrule
        Type & Method & NLL \ $\downarrow$ & AS (\%)\ $\uparrow$ & MS (\%)\ $\uparrow$ & Val (\%)\ $\uparrow$ & Val and Uniq (\%)\ $\uparrow$ \\ \midrule
        NF & E-NF & -59.7 & 85.0 & 4.9 & 40.2 & 39.4 \\ \midrule
        Generative GNN & G-Schnet & - & 95.7 & 68.1 & 85.5 & 80.3 \\ \midrule
        DDPM & GDM & -94.7 & 97.0 & 63.2 & - & - \\
        & GDM-aug & -92.5 & 97.6 & 71.6 & 90.4 & 89.5 \\
        & EDM & -110.7 $\pm$ 1.5 & 98.7 $\pm$ 0.1 & 82.0 $\pm$ 0.4 & 91.9 $\pm$ 0.5 & 90.7 $\pm$ 0.6 \\
        & Bridge & - & 98.7 $\pm$ 0.1 & 81.8 $\pm$ 0.2 & - & 90.2 \\
        & Bridge + Force & - & 98.8 $\pm$ 0.1 & 84.6 $\pm$ 0.3 & 92.0 & 90.7 \\ \midrule
        LDM & GraphLDM & - & 97.2 & 70.5 & 83.6 & 82.7 \\
        & GraphLDM-aug & - & 97.9 & 78.7 & 90.5 & 89.5 \\
        & GeoLDM & - & \textbf{98.9} $\pm$ 0.1 & \textbf{89.4} $\pm$ 0.5 & 93.8 $\pm$ 0.4 & 92.7 $\pm$ 0.5 \\ \midrule
        GC-DDPM - \textit{Ours} & \textsc{GCDM} w/o Frames & \underline{-162.3} $\pm$ 0.3 & 98.4 $\pm$ 0.0 & 81.7 $\pm$ 0.5 & \underline{93.9} $\pm$ 0.1 & \underline{92.7} $\pm$ 0.1 \\
        & \textsc{GCDM} w/o SMA & -131.3 $\pm$ 0.8 & 95.7 $\pm$ 0.1 & 51.7 $\pm$ 1.4 & 83.1 $\pm$ 1.7 & 82.8 $\pm$ 1.7 \\
        \rowcolor[gray]{0.8} & \textsc{GCDM} & \textbf{-171.0} $\pm$ 0.2 & \underline{98.7} $\pm$ 0.0 & \underline{85.7} $\pm$ 0.4 & \textbf{94.8} $\pm$ 0.2 & \textbf{93.3} $\pm$ 0.0 \\ \midrule
        Data & & & 99.0 & 95.2 & 97.7 & 97.7 \\ \midrule
    \end{tabular}%
}
\\
\resizebox{\linewidth}{!}{%
    \begin{tabular}{llllllll}
        & & & & \\ \midrule
        Method & NLL \ $\downarrow$ & AS (\%)\ $\uparrow$ & MS (\%)\ $\uparrow$ & Val (\%)\ $\uparrow$ & Val and Uniq (\%)\ $\uparrow$ & Novel (\%)\ $\uparrow$ & PB-Valid (\%)\ $\uparrow$ \\ \midrule
        GeoLDM & - & \textbf{98.9} $\pm$ 0.0 & \textbf{89.8} $\pm$ 0.4 & \underline{93.6} $\pm$ 0.2 & \underline{91.8} $\pm$ 0.2 & \underline{53.5} $\pm$ 0.6 & \textbf{93.1} $\pm$ 0.4 \\ \midrule
        \rowcolor[gray]{0.8} \textsc{GCDM} & \textbf{-169.4} $\pm$ 0.8 & \underline{98.7} $\pm$ 0.1 & \underline{86.0} $\pm$ 0.7 & \textbf{94.9} $\pm$ 0.3 & \textbf{93.4} $\pm$ 0.3 & \textbf{58.7} $\pm$ 0.5 & \underline{91.9} $\pm$ 0.5 \\ \midrule
    \end{tabular}
}
\caption{Comparison of \textsc{GCDM} with baseline methods for 3D molecule generation. The results in the top half of the table are reported in terms of the negative log-likelihood (NLL) - $\log p(\mathbf{x}, \mathbf{h}, N)$, atom stability, molecule stability, validity, and uniqueness of 10,000 samples drawn from each model, with standard deviations ($\pm$) for each model across three runs on QM9. The results in the bottom half of the table are for methods specifically evaluated across \textit{five} runs on QM9 using Student's t-distribution 95\% confidence intervals for per-metric errors, additionally with novelty (Novel) defined as the percentage of (valid and unique) generated molecule SMILES strings that were not found in the QM9 dataset and PoseBusters validity (PB-Valid) defined as the percentage of generated molecules that pass all relevant \textit{de novo} structural and chemical sanity checks listed in Section \ref{section:unconditional_qm9}. The top-1 (best) results for this task are in \textbf{bold}, and the second-best results are \underline{underlined}, with - denoting a metric value that is not available.}
\label{table:3dmg_unconditional_qm9_results}
\end{table}

\begin{figure}[hbt!]
  \centering
  \subfloat[\textrm{[H]/...}]{\fbox{\includegraphics[height=2.3cm]{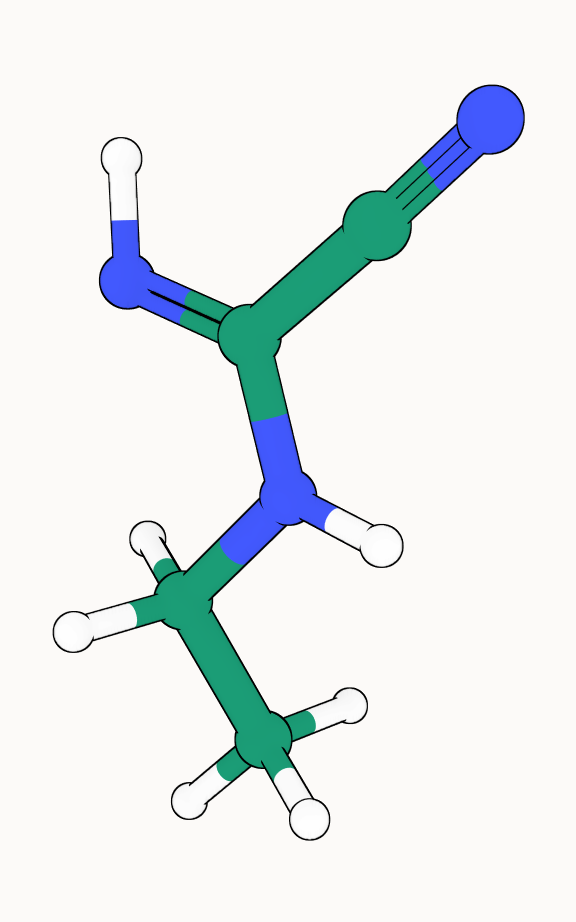}}}
  \subfloat[\textrm{CC[N...}]{\fbox{\includegraphics[height=2.3cm]{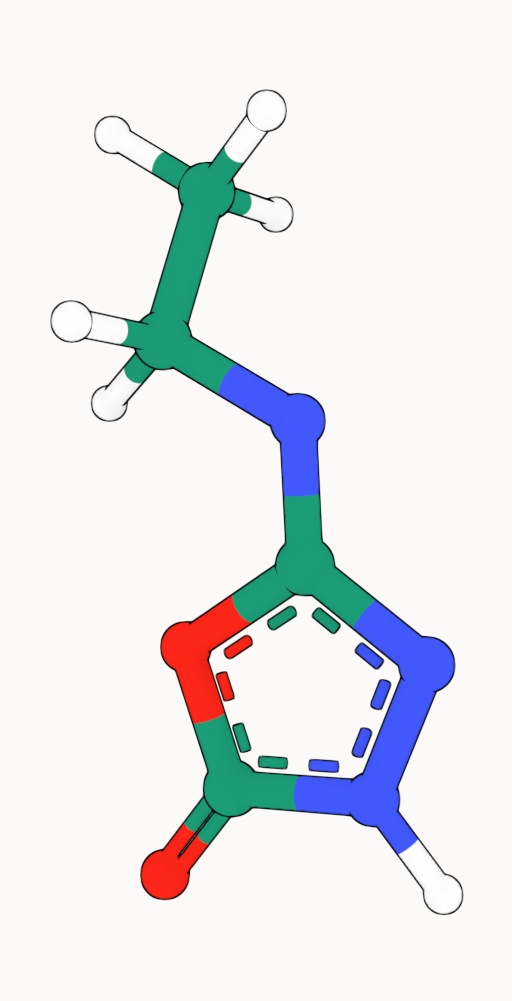}}}
  \subfloat[\textrm{O=CC...}]{\fbox{\includegraphics[height=2.3cm]{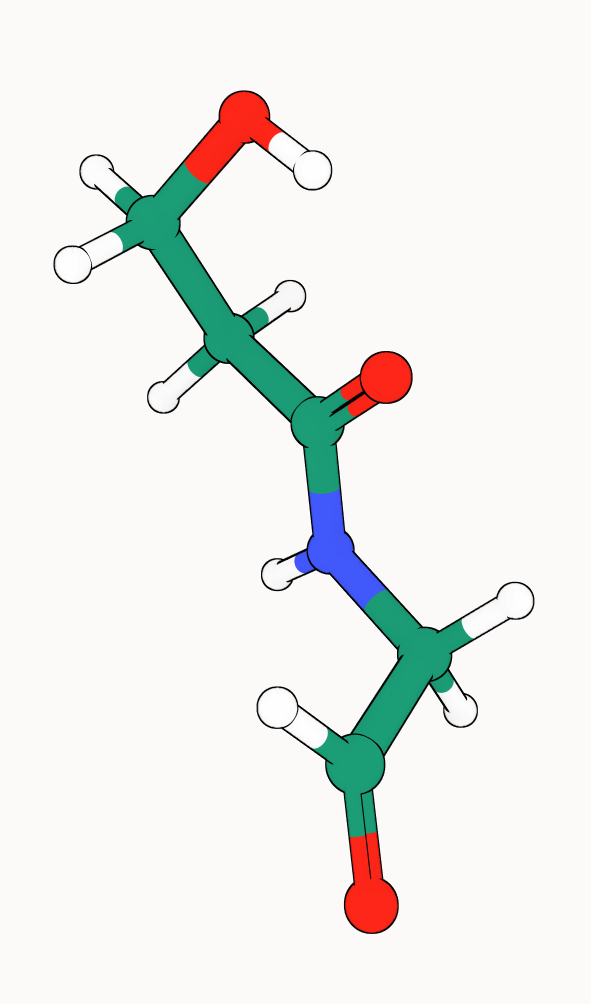}}}
  \subfloat[\textrm{C/N=...}]{\fbox{\includegraphics[height=2.3cm]{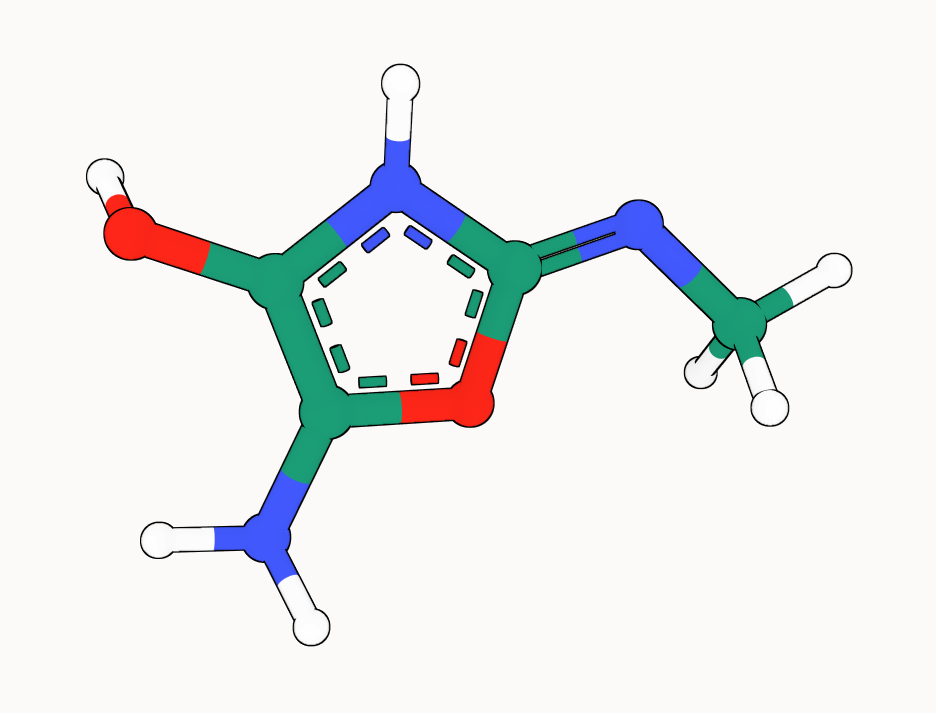}}}
  \subfloat[\textrm{[H]/...}]{\fbox{\includegraphics[height=2.3cm]{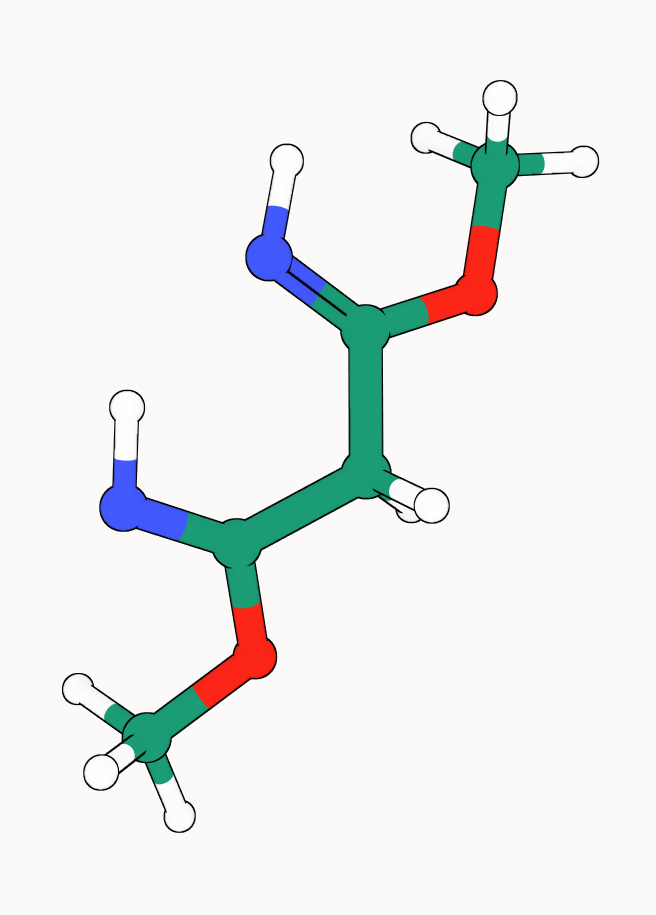}}}
  \subfloat[\textrm{Oc1c...}]{\fbox{\includegraphics[height=2.3cm]{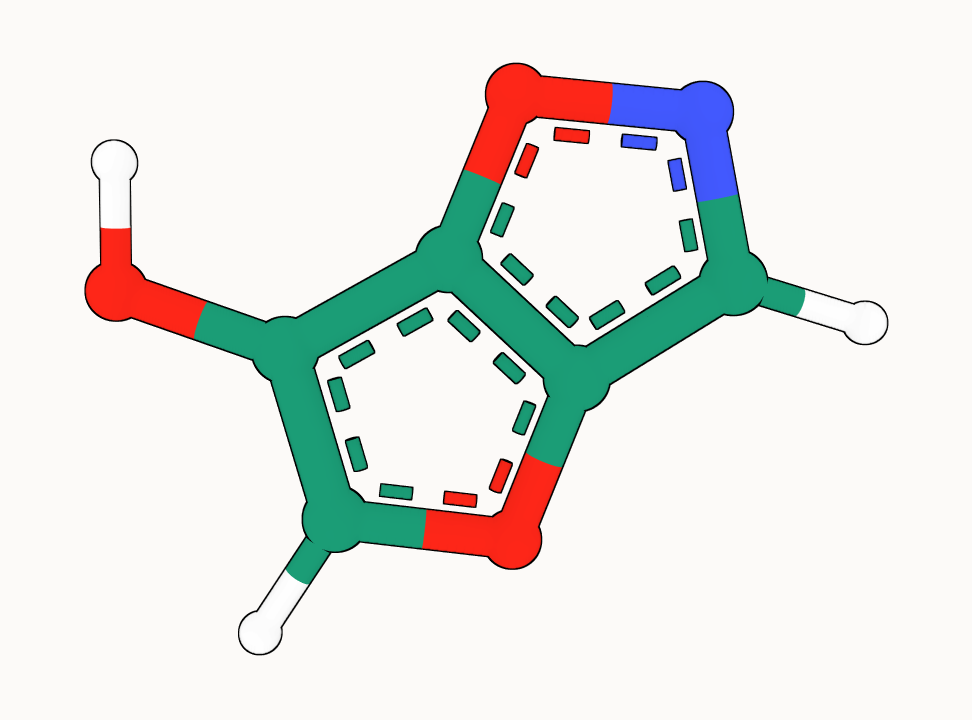}}}
  \caption{PB-valid 3D molecules generated by \textsc{GCDM} for the QM9 dataset.}
  \label{fig:unconditional_qm9_samples}
\end{figure}

\textbf{Results.} In the top half of Table \ref{table:3dmg_unconditional_qm9_results}, we see that \textsc{GCDM} achieves the highest percentage of probable (NLL), valid, and unique molecules compared to all baseline methods, with AS and MS results marginally lower than those of GeoLDM yet with lower standard deviations. In the bottom half of Table \ref{table:3dmg_unconditional_qm9_results}, , where we reevaluate \textsc{GCDM} and GeoLDM using 5 sampling runs and report 95\% confidence intervals for each metric, \textsc{GCDM} generates 1.6\% more RDKit-valid and unique molecules and 5.2\% more novel molecules compared to GeoLDM, all while offering the best reported negative log-likelihood (NLL) for the QM9 test dataset. This result indicates that although GeoLDM offers novelty rates close to parity (i.e., 50\%), \textsc{GCDM} nearly matches the stability and PB-validity rates of GeoLDM while yielding novel molecules nearly 60\% of the time on average, suggesting that \textsc{GCDM} may prove more useful for accurately exploring the space of novel yet valid small molecules. Our ablation of SMA within \textsc{GCDM} demonstrates that, to generate stable 3D molecules, \textsc{GCDM} heavily relies on both being able to perform a lightweight version of fully-connected graph self-attention \citep{vaswani2017attention}, which suggests avenues of future research that will be required to scale up such generative models to large biomolecules such as proteins. Additionally, removing geometric local frame embeddings from \textsc{GCDM} reveals that the inductive biases of molecular chirality and geometry-completeness are important contributing factors in \textsc{GCDM} achieving these SOTA results. Figure \ref{fig:unconditional_qm9_samples} illustrates PoseBusters-valid examples of QM9-sized molecules generated by \textsc{GCDM}, with the following corresponding SMILES strings from left to right:
\textbf{(\underline{a})} \textrm{[H]/N=C(\textbackslash C\#N)NCC},
\textbf{(\underline{b})} \textrm{CC[N]c1n[nH]c(=O)o1},
\textbf{(\underline{c})} \textrm{O=CCNC(=O)CCO},
\textbf{(\underline{d})} \textrm{C/N=c1/[nH]c(O)c(N)o1},
\textbf{(\underline{e})} \textrm{[H]/N=C(/C[C]([NH])OC)OC}, and
\textbf{(\underline{f})} \textrm{Oc1coc2cnoc12}.

\subsection{Property-Conditional 3D Molecule Generation - QM9}
\label{section:conditional_qm9}

\textbf{Baselines.} Towards the practical use case of conditional generation of 3D molecules, we compare \textsc{GCDM} to existing E(3)-equivariant models, EDM \citep{hoogeboom2022equivariant} and GeoLDM \citep{xu2023geometric}, as well as to two naive baselines: "Naive (Upper-bound)" where a molecular property classifier $\phi_{c}$ predicts molecular properties given a method's generated 3D molecules and shuffled (i.e., random) property labels; and "\# Atoms" where one uses the numbers of atoms in a method's generated 3D molecules to predict their molecular properties. For each baseline method, we report its mean absolute error (MAE) in terms of molecular property prediction by an ensemble of three EGNN classifiers $\phi_{c}$ \citep{satorras2021n} as reported in \citet{hoogeboom2022equivariant}. For \textsc{GCDM}, we train each conditional model by conditioning it on one of six distinct molecular property feature inputs - $\alpha$, gap, homo, lumo, $\mu$, and $C_{v}$ - for approximately 1,500 epochs using the QM9 validation split of \citet{hoogeboom2022equivariant} as the model's training dataset and the QM9 training split of \citet{hoogeboom2022equivariant} as the corresponding EGNN classifier ensemble's training dataset. Consequently, one can expect the gap between a method's performance and that of "QM9 (Lower-bound)" to decrease as the method more accurately generates property-specific molecules.

\begin{table}[t]
\centering
\resizebox{\textwidth}{!}{%
    \begin{tabular}{lllllll}
        & & & & & & \\ \midrule
        Task & $\alpha$\ $\downarrow$ & $\Delta \epsilon$\ $\downarrow$ & $\epsilon_{HOMO}$\ $\downarrow$ & $\epsilon_{LUMO}$\ $\downarrow$ & $\mu$\ $\downarrow$ & $C_{v}$\ $\downarrow$ \\
        Units & $Bohr^{3}$ & $meV$ & $meV$ & $meV$ & $D$ & $\frac{cal}{mol} K$ \\ \midrule
        Naive (Upper-bound) & 9.01 & 1470 & 645 & 1457 & 1.616 & 6.857 \\
        \# Atoms & 3.86 & 866 & 426 & 813 & 1.053 & 1.971 \\
        EDM & 2.76 & 655 & 356 & 584 & 1.111 & 1.101 \\
        GeoLDM & \underline{2.37} & \textbf{587} & \textbf{340} & \underline{522} & \underline{1.108} & \underline{1.025} \\
        \rowcolor[gray]{0.8} \textsc{GCDM} & \textbf{1.97} & \underline{602} & \underline{344} & \textbf{479} & \textbf{0.844} & \textbf{0.689} \\ \midrule
        QM9 (Lower-bound) & 0.10 & 64 & 39 & 36 & 0.043 & 0.040 \\ \midrule
    \end{tabular}
}%
\\
\resizebox{\textwidth}{!}{%
    \begin{tabular}{lllllll}
        & & & & & & \\ \midrule
        Task & $\alpha$\ $\downarrow$ & $\Delta \epsilon$\ $\downarrow$ & $\epsilon_{HOMO}$\ $\downarrow$ & $\epsilon_{LUMO}$\ $\downarrow$ & $\mu$\ $\downarrow$ & $C_{v}$\ $\downarrow$ \\
        Units & $Bohr^{3}$ & $meV$ & $meV$ & $meV$ & $D$ & $\frac{cal}{mol} K$ \\ \midrule
        GeoLDM & 2.77 $\pm$ 0.12 & \underline{655} $\pm$ 20.57 & 357 $\pm$ 5.68 & \underline{565} $\pm$ 10.62 & \underline{1.089} $\pm$ 0.02 & 1.070 $\pm$ 0.04 \\
        \rowcolor[gray]{0.8} \textsc{GCDM} & \textbf{1.99} $\pm$ 0.01 & \textbf{595} $\pm$ 14.34 & \textbf{346} $\pm$ 1.23 & \textbf{480} $\pm$ 6.58 & \textbf{0.855} $\pm$ 0.00 & \textbf{0.698} $\pm$ 0.01 \\ \midrule
    \end{tabular}%
}
\resizebox{\textwidth}{!}{%
    \begin{tabular}{lllllll}
        Metric & $\alpha$\ PB-Valid (\%)\ $\uparrow$ & $\Delta \epsilon$\ PB-Valid (\%)\ $\uparrow$ & $\epsilon_{HOMO}$\ PB-Valid (\%)\ $\uparrow$ & $\epsilon_{LUMO}$\ PB-Valid (\%)\ $\uparrow$ & $\mu$\ PB-Valid (\%)\ $\uparrow$ & $C_{v}$\ PB-Valid (\%)\ $\uparrow$ \\ \midrule
        GeoLDM & 93.7 $\pm$ 0.5 & 92.8 $\pm$ 0.3 & 93.9 $\pm$ 0.4 & 93.3 $\pm$ 0.6 & 93.2 $\pm$ 1.3 & 92.5 $\pm$ 0.8 \\
        \rowcolor[gray]{0.8} \textsc{GCDM} & 92.3 $\pm$ 0.3 & 92.5 $\pm$ 0.8 & 92.7 $\pm$ 0.5 & 92.7 $\pm$ 0.6 & 92.4 $\pm$ 0.4 & 91.7 $\pm$ 0.4 \\ \midrule
    \end{tabular}%
}
\caption{Comparison of \textsc{GCDM} with baseline methods for property-conditional 3D molecule generation. The results in the top half of the table are reported in terms of the MAE for molecular property prediction by an EGNN classifier $\phi_{c}$ on a QM9 subset, with results listed for \textsc{GCDM}-generated samples as well as for four separate baseline methods. The results in the bottom half of the table (where GeoLDM is retrained using its official code repository due to the unavailability of its conditional model checkpoints) are likewise listed for selected methods yet instead report (across an ensemble of three separately-trained EGNN property classifier models, each with a distinct random seed) Student's t-distribution 95\% confidence error intervals for each property metric as well as the percentage of PoseBusters-validated (PB-Valid) \textit{de novo} generated molecules. The top-1 (best) conditioning results for this task are in \textbf{bold}, and the second-best results are \underline{underlined}.}
\label{table:3dmg_conditional_qm9_results}
\end{table}

\begin{figure}[hbt!]
  \centering
  
  \subfloat[$\alpha=68.7$]{\fbox{\includegraphics[height=1.8cm]{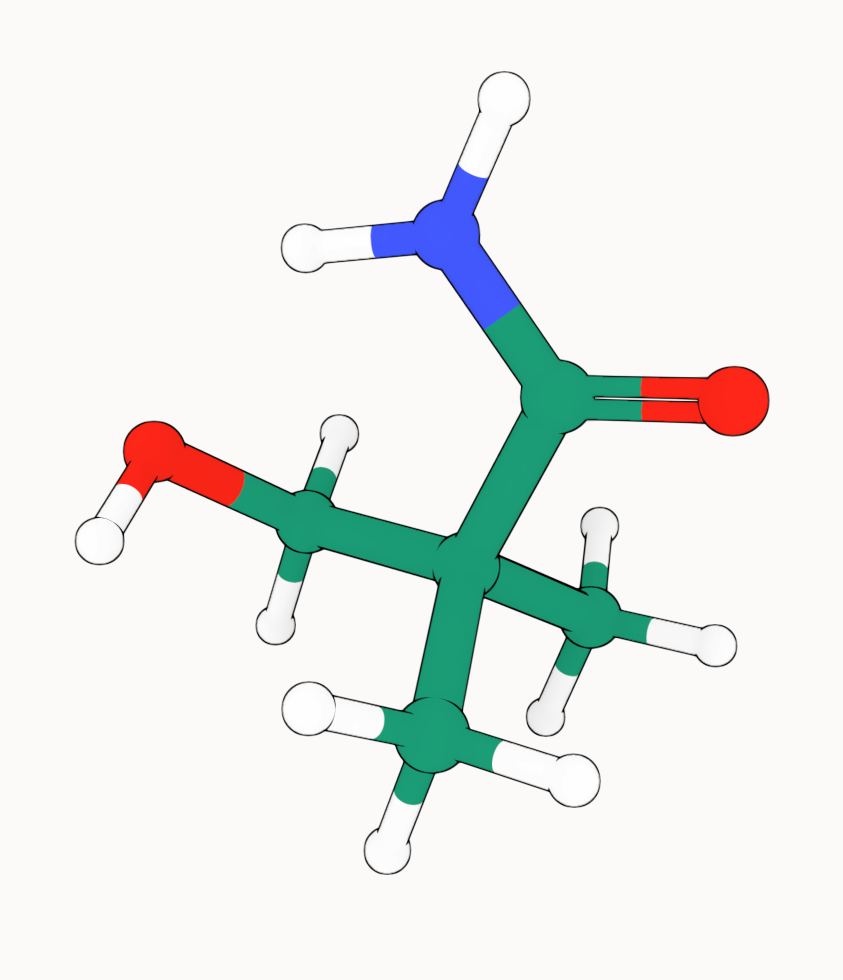}}}
  \subfloat[$73.5$]{\fbox{\includegraphics[height=1.8cm]{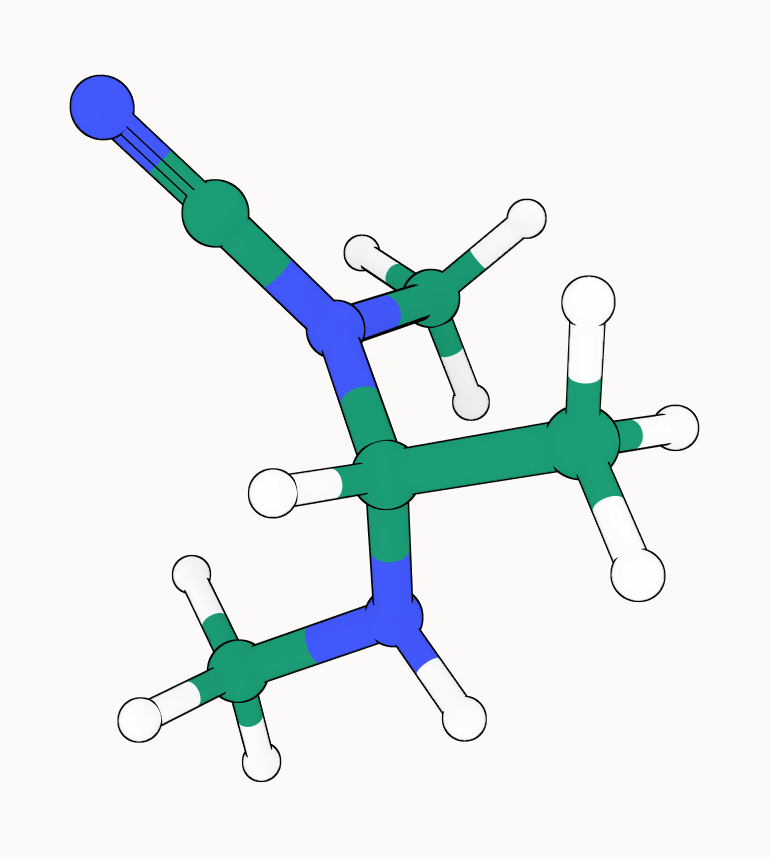}}}
  \subfloat[$79.1$]{\fbox{\includegraphics[height=1.8cm]{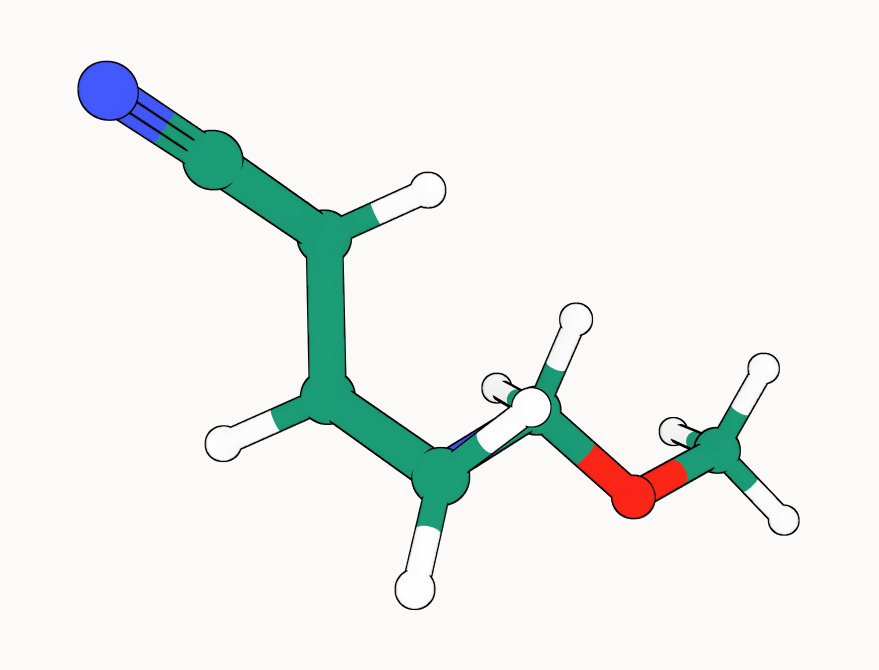}}}
  \subfloat[$82.5$]{\fbox{\includegraphics[height=1.8cm]{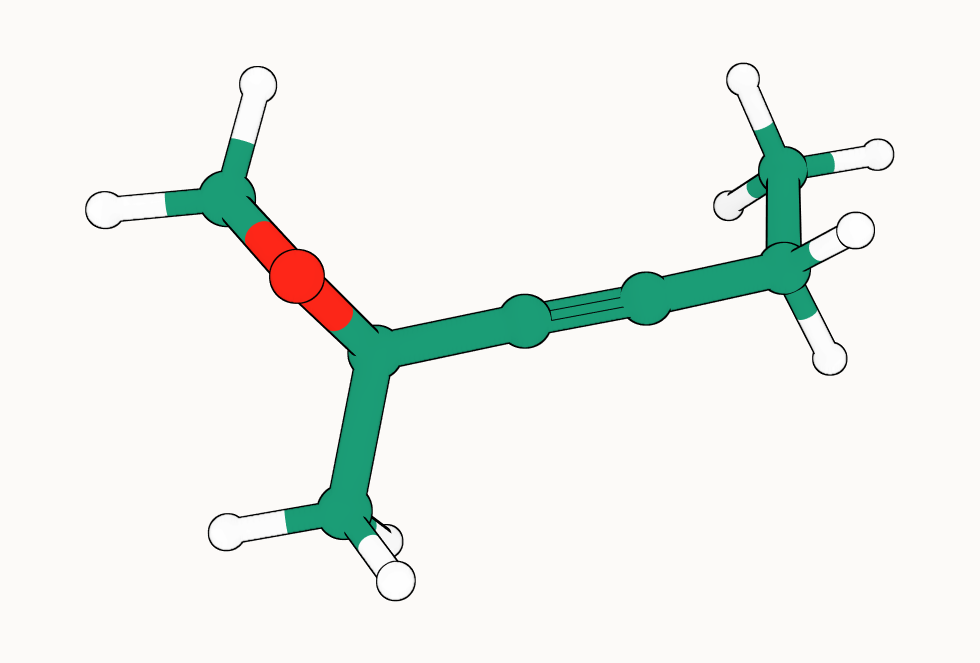}}}
  \subfloat[$86.0$]{\fbox{\includegraphics[height=1.8cm]{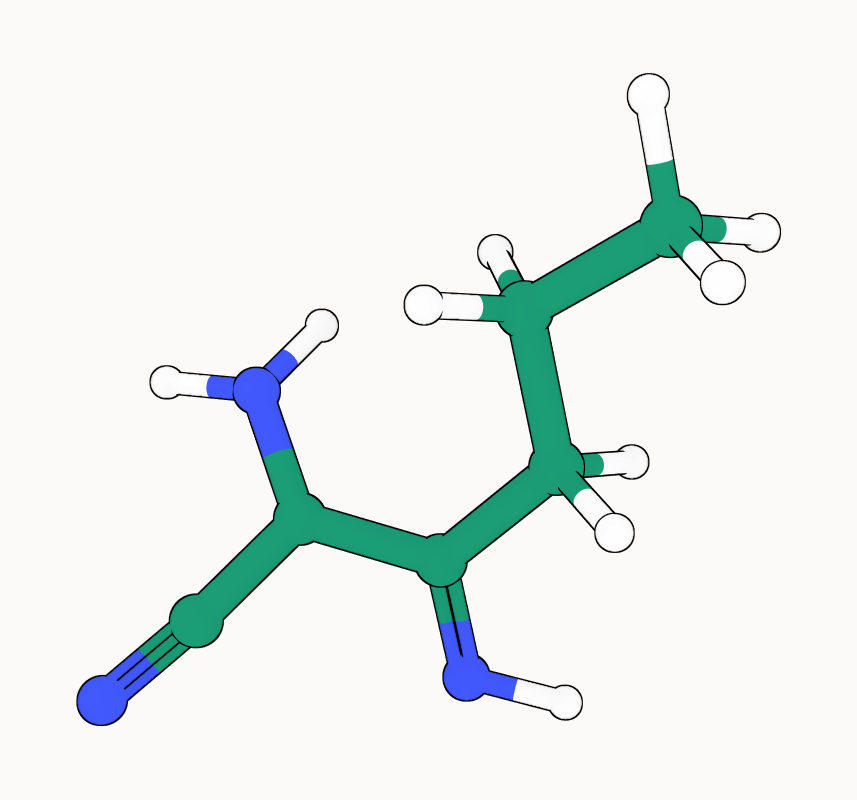}}}
  \subfloat[$93.6$]{\fbox{\includegraphics[height=1.8cm]{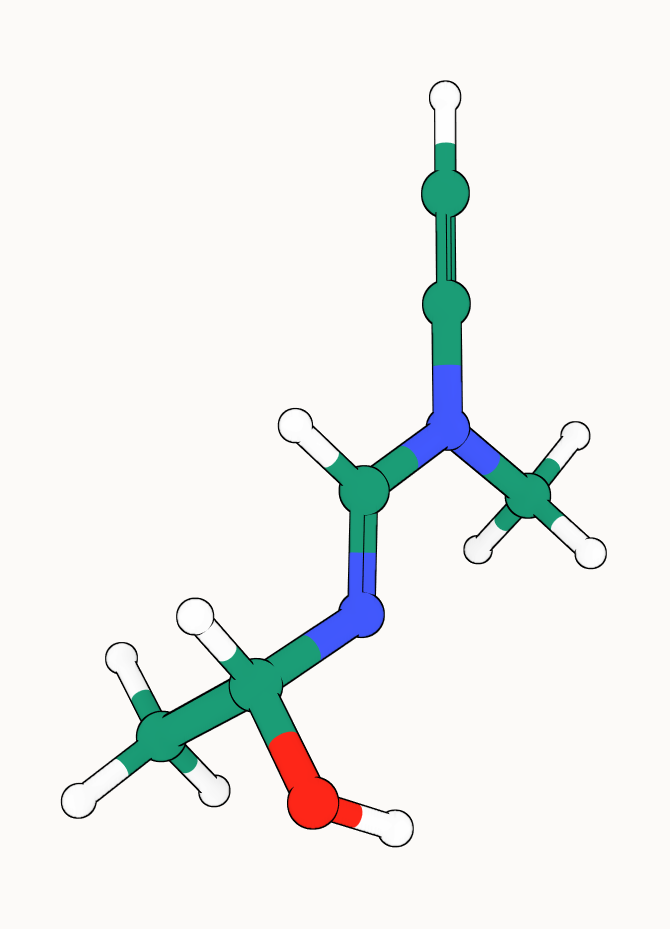}}}
  
  \caption{PB-valid 3D molecules generated by \textsc{GCDM} using increasing values of $\alpha$.}
  \label{fig:conditional_qm9_samples}
\end{figure}

\textbf{Results.} We see in Table \ref{table:3dmg_conditional_qm9_results} that \textsc{GCDM} achieves the best overall results compared to all baseline methods in conditioning on a given molecular property, with conditionally-generated samples shown in Figure \ref{fig:conditional_qm9_samples} (Note: \textsc{PSI4}-computed property values \citep{smith2020psi4} for (a) and (f) are 69.1 Bohr$^{3}$ (energy: -402 a.u.) and 89.7 Bohr$^{3}$ (energy: -419 a.u.), respectively, at the \textsc{DFT/B3LYP/6-31G(2df,p)} level of theory \citep{ramakrishnan2014quantum, lehtola2018recent}). In particular, as shown in the bottom half of this table, \textsc{GCDM} surpasses the MAE results of the SOTA GeoLDM method (by 19\% on average) for all six molecular properties - $\alpha$, gap, homo, lumo, $\mu$, and $C_{v}$ - by 28\%, 9\%, 3\%, 15\%, 21\%, and 35\%, respectively, while nearly matching the PB-Valid rates of GeoLDM (similar to the results in Table \ref{table:3dmg_unconditional_qm9_results}). These results qualitatively and quantitatively demonstrate that, using geometry-complete diffusion, \textsc{GCDM} enables notably precise generation of 3D molecules with specific molecular properties (e.g., $\alpha$ - polarizability).

\subsection{Unconditional 3D Molecule Generation - GEOM-Drugs}
\label{section:unconditional_geom}
The second dataset used in our experiments, the GEOM-Drugs dataset, is a well-known source of large, 3D molecular conformers for downstream machine learning tasks. It contains 430k molecules, each with 44 atoms on average and with up to as many as 181 atoms after hydrogen atoms are imputed for each molecule following dataset postprocessing as in \citet{hoogeboom2022equivariant}. For this experiment, we collect the 30 lowest-energy conformers corresponding to a molecule and task each baseline method with generating new molecules with 3D positions and types for each constituent atom. Here, we also adopt the negative log-likelihood, atom stability, and molecule stability metrics as defined in Section \ref{section:unconditional_qm9} and train \textsc{GCDM} using the same hyperparameters as listed in Appendix \ref{section:appendix_training_details}, with the exception of training for approximately 75 epochs on GEOM-Drugs.

\begin{table}[t]
\centering
\resizebox{\textwidth}{!}{%
    \begin{tabular}{lllll}
        & & & & \\ \midrule
        Type & Method & NLL $\downarrow$ & AS (\%)\ $\uparrow$ & MS (\%)\ $\uparrow$ \\ \midrule
        NF & E-NF & - & 75.0 & 0.0 \\ \midrule
        DDPM & GDM & -14.2 & 75.0 & 0.0 \\
        & GDM-aug & -58.3 & 77.7 & 0.0 \\
        & EDM & \underline{-137.1} & 81.3 & 0.0 \\
        & Bridge & - & 81.0 $\pm$ 0.7 & 0.0 \\
        & Bridge + Force & - & 82.4 $\pm$ 0.8 & 0.0 \\ \midrule
        LDM & GraphLDM & - & 76.2 & 0.0 \\
        & GraphLDM-aug & - & 79.6 & 0.0 \\
        & GeoLDM & - & \underline{84.4} & \underline{0.0} \\ \midrule
        GC-DDPM - \textit{Ours} & \textsc{GCDM} w/o Frames & 769.7 & 88.0 $\pm$ 0.3 & 3.4 $\pm$ 0.3 \\
        & \textsc{GCDM} w/o SMA & 3505.5 & 43.9 $\pm$ 3.6 & 0.1 $\pm$ 0.0 \\
        \rowcolor[gray]{0.8} & \textsc{GCDM} & \textbf{-234.3} & \textbf{89.0} $\pm$ 0.8 & \textbf{5.2} $\pm$ 1.1 \\ \midrule
        Data & & & 86.5 & 2.8 \\ \midrule
    \end{tabular}%
}
\\
\resizebox{\linewidth}{!}{%
    \begin{tabular}{llllllll}
        & & & & \\ \midrule
        Method & NLL \ $\downarrow$ & AS (\%)\ $\uparrow$ & MS (\%)\ $\uparrow$ & Val (\%)\ $\uparrow$ & Val and Uniq (\%)\ $\uparrow$ & Novel (\%)\ $\uparrow$ & PB-Valid (\%)\ $\uparrow$ \\ \midrule
        GeoLDM & - & \underline{84.4} $\pm$ 0.1  & \underline{0.6} $\pm$ 0.1 & \textbf{99.5} $\pm 0.1$ & \textbf{99.4} $\pm$ 0.1 & - & \underline{38.3} $\pm$ 0.5 \\ \midrule
        \rowcolor[gray]{0.8} \textsc{GCDM} & \textbf{-215.1} $\pm$ 3.8 & \textbf{88.1} $\pm$ 0.1 & \textbf{4.3} $\pm$ 0.4 & \underline{95.5} $\pm$ 0.1 & \underline{95.5} $\pm$ 0.1 & \textbf{95.5} $\pm$ 0.1 & \textbf{77.0} $\pm$ 0.1 \\ \midrule
    \end{tabular}
}
\caption{Comparison of \textsc{GCDM} with baseline methods for 3D molecule generation. The results in the top half of the table are reported in terms of each method's negative log-likelihood, atom stability, and molecule stability with standard deviations ($\pm$) across three runs on GEOM-Drugs, each drawing 10,000 samples from the model. The results in the bottom half of the table are for methods specifically evaluated across \textit{five} runs on QM9 using Student's t-distribution 95\% confidence intervals for per-metric errors, additionally with validity and uniqueness (Val and Uniq), novelty (Novel), and PoseBusters validity (PB-Valid) defined likewise as in Section \ref{section:unconditional_qm9}; The top-1 (best) results for this task are in \textbf{bold}, and the second-best results are \underline{underlined}.}
\label{table:3dmg_unconditional_geom_results}
\end{table}

\begin{figure}[hbt!]
  \centering
  
  \subfloat[\textrm{CC(C)...}]{\fbox{\includegraphics[height=2.3cm]{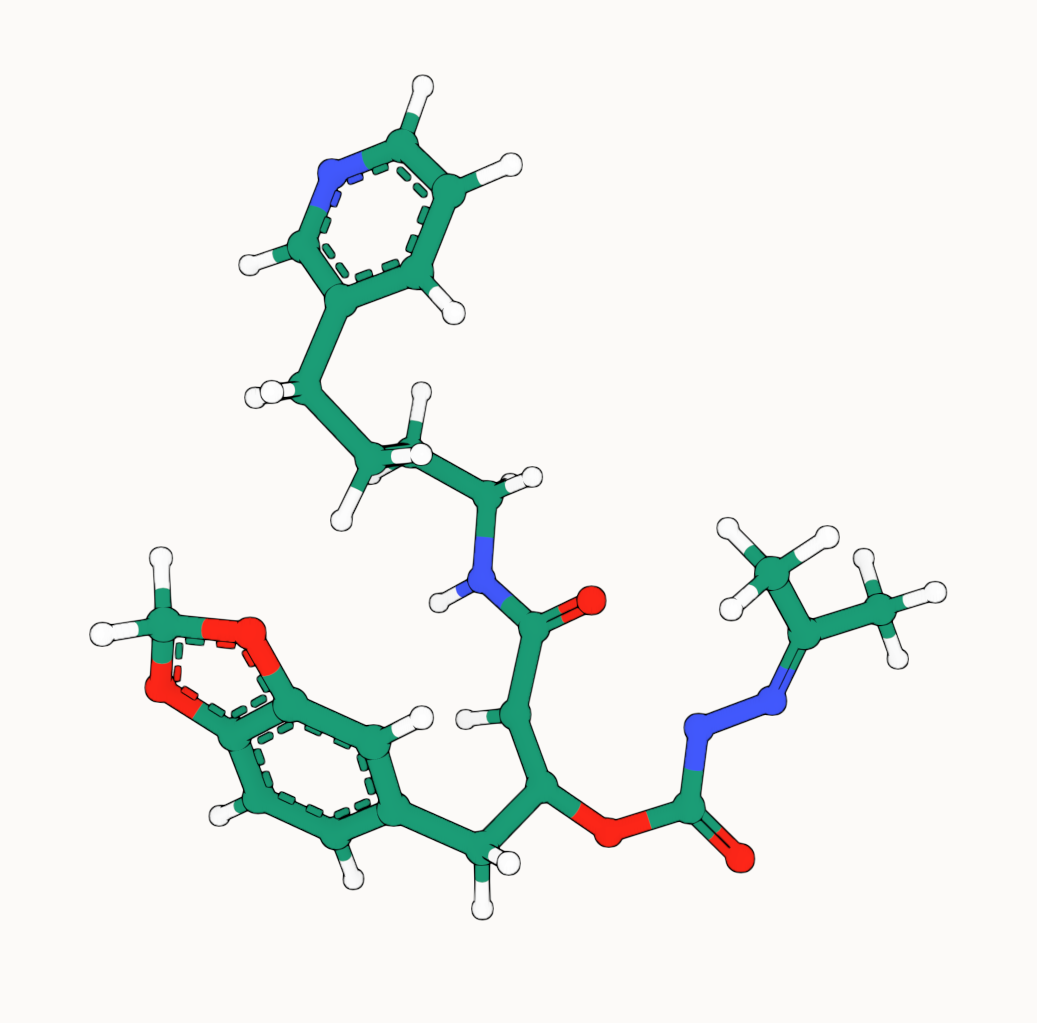}}}
  \subfloat[\textrm{CN(N)...}]{\fbox{\includegraphics[height=2.3cm]{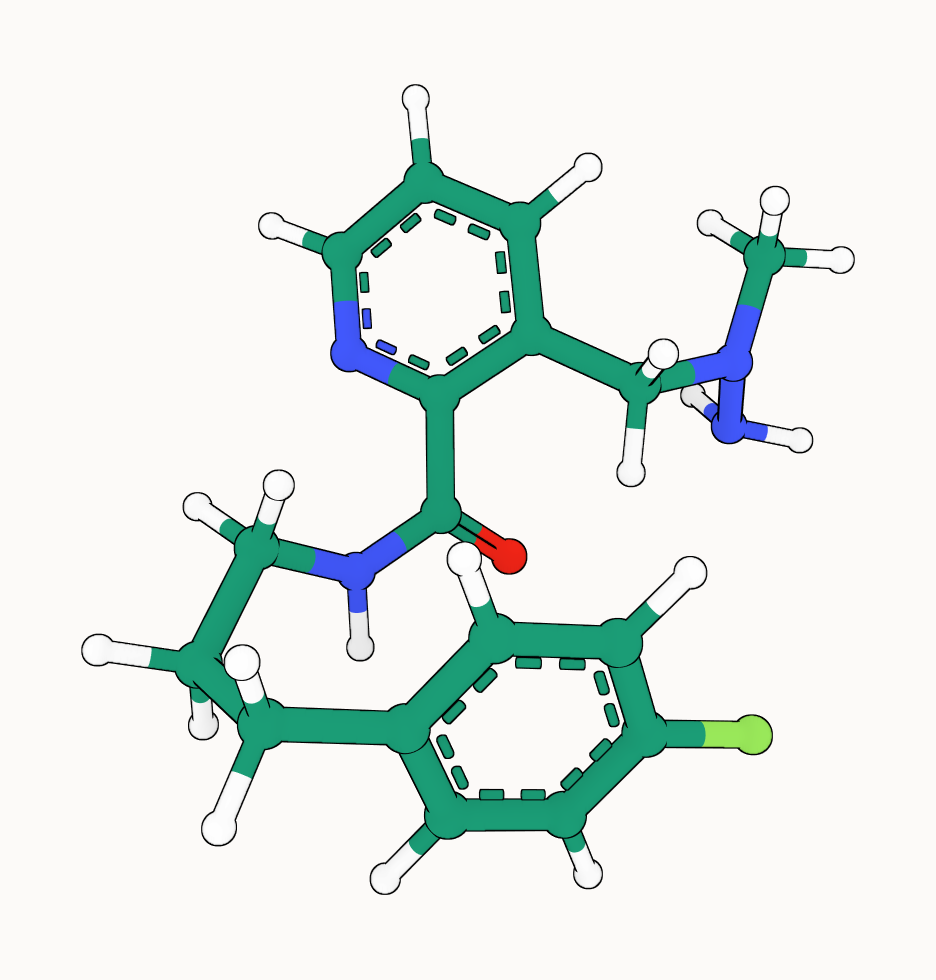}}}
  \subfloat[\textrm{C=CCC...}]{\fbox{\includegraphics[height=2.3cm]{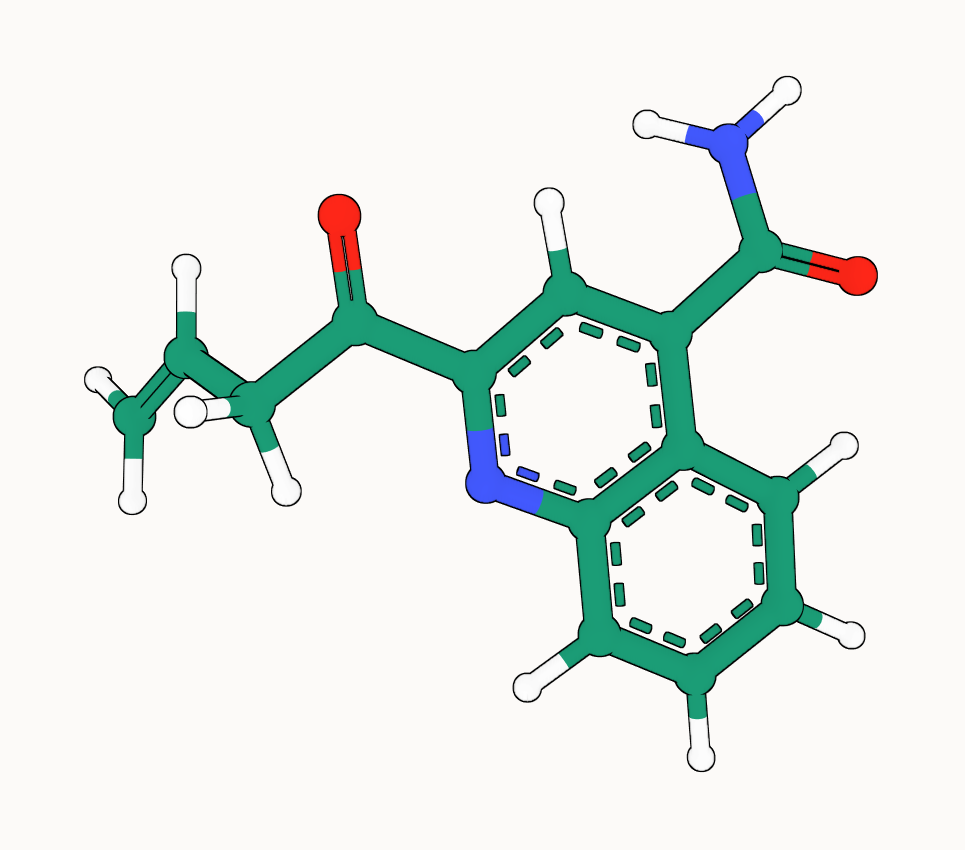}}}
  \subfloat[\textrm{CC(==...}]{\fbox{\includegraphics[height=2.3cm]{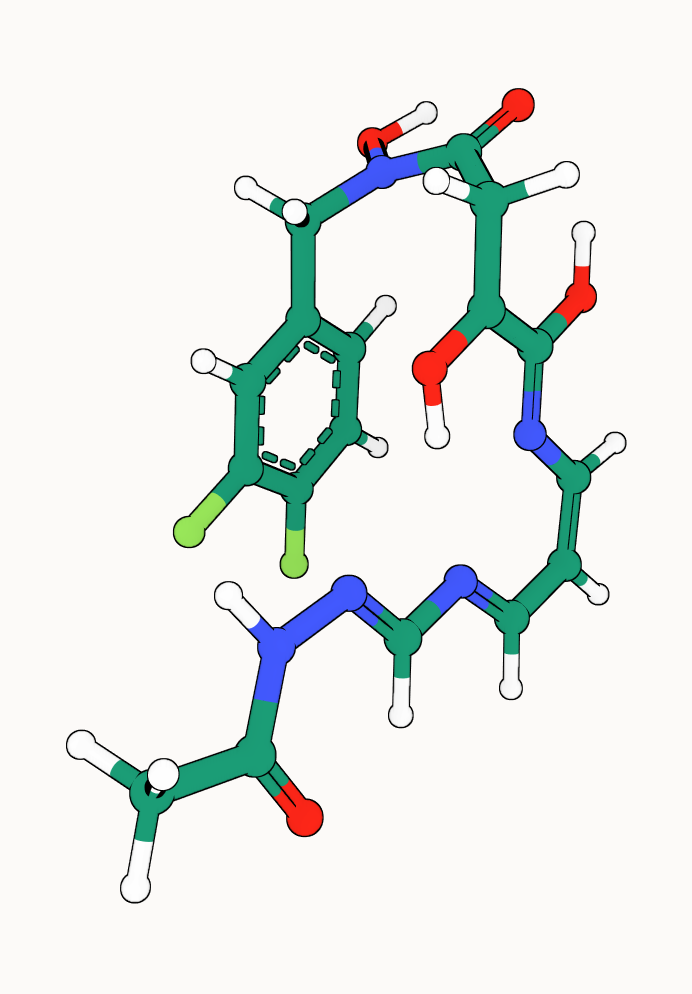}}}
  \subfloat[\textrm{COC(=...}]{\fbox{\includegraphics[height=2.3cm]{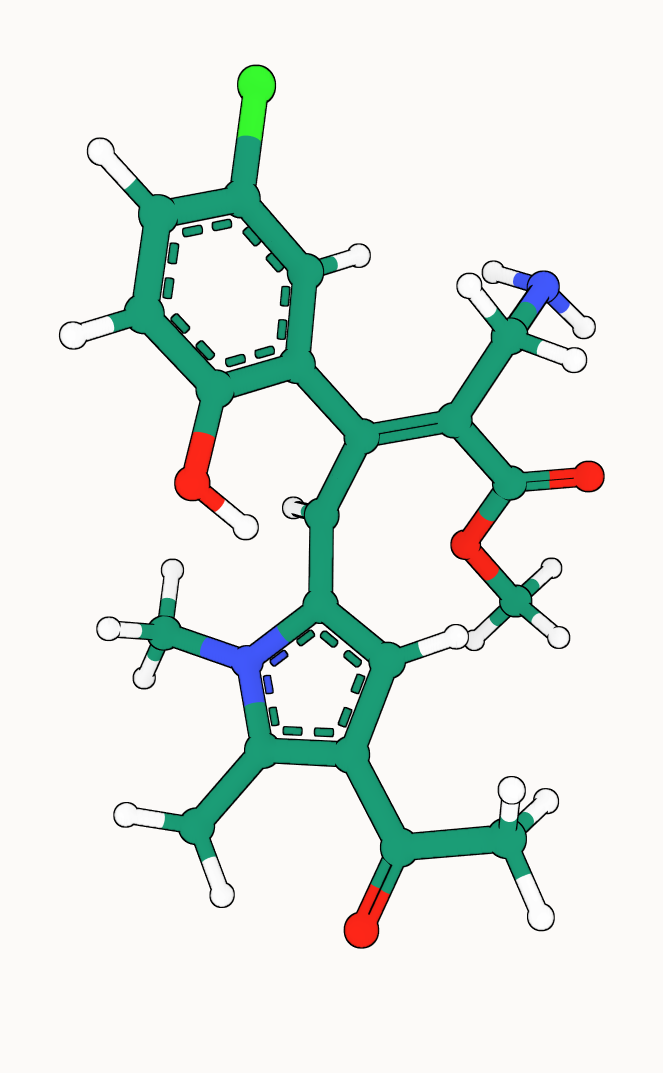}}}
  \subfloat[\textrm{CC[C@...}]{\fbox{\includegraphics[height=2.3cm]{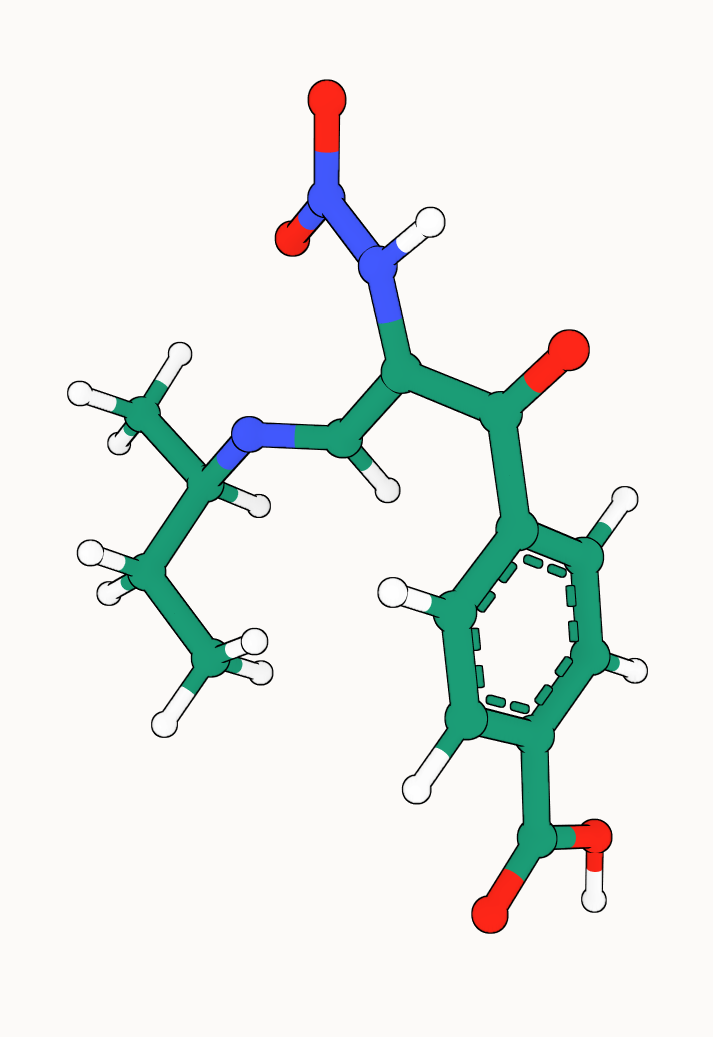}}}
  
  \caption{PB-valid 3D molecules generated by \textsc{GCDM} for the GEOM-Drugs dataset.}
  \label{fig:unconditional_geom_samples}
\end{figure}

\textbf{Baselines.} In this experiment, we compare \textsc{GCDM} to several state-of-the-art baseline methods for 3D molecule generation on GEOM-Drugs. Similar to our experiments on QM9, in addition to including a reference point for molecule quality metrics using GEOM-Drugs itself (i.e., Data), here we also compare against E-NF, GDM, GDM-aug, EDM, Bridge along with its variant Bridge + Force, as well as GraphLDM, GraphLDM-aug, and GeoLDM. As in Section \ref{section:unconditional_qm9} , each method's results in the top half (bottom half) of the table are reported as the mean and standard deviation (mean and Student's t-distribution 95\% confidence interval) ($\pm$) of each metric across three (five) test runs on GEOM-Drugs.

\textbf{Results.} To start, Table \ref{table:3dmg_unconditional_geom_results} displays an interesting phenomenon that is important to note: Due to the size and atomic complexity of GEOM-Drugs' molecules and the subsequent errors accumulated when estimating bond types based on such inter-atom distances, the baseline results for the molecule stability metrics measured here (i.e., Data) are much lower than those collected for the QM9 dataset. Thus, reporting additional chemical and structural validity metrics (e.g., PB-Valid) for comparison is crucial to accurately assess a method's performance in this context, which we do in the bottom half of Table \ref{table:3dmg_unconditional_geom_results}. Nonetheless, for GEOM-Drugs, \textsc{GCDM} supersedes EDM's SOTA negative log-likelihood results by 57\% and advances GeoLDM's SOTA atom and molecule stability results by 4\% and more than sixfold, respectively. More importantly, however, \textsc{GCDM} can generate a significant proportion of PB-valid large molecules, surpassing even the reference molecule stability rate of the GEOM-Drugs dataset (i.e., 2.8) by 54\%, demonstrating that geometric diffusion models such as \textsc{GCDM} can not only effectively generate valid large molecules but can also generalize beyond the native distribution of stable molecules within GEOM-Drugs.

\begin{figure}[t]
\centering
\includegraphics[width=0.75\textwidth]{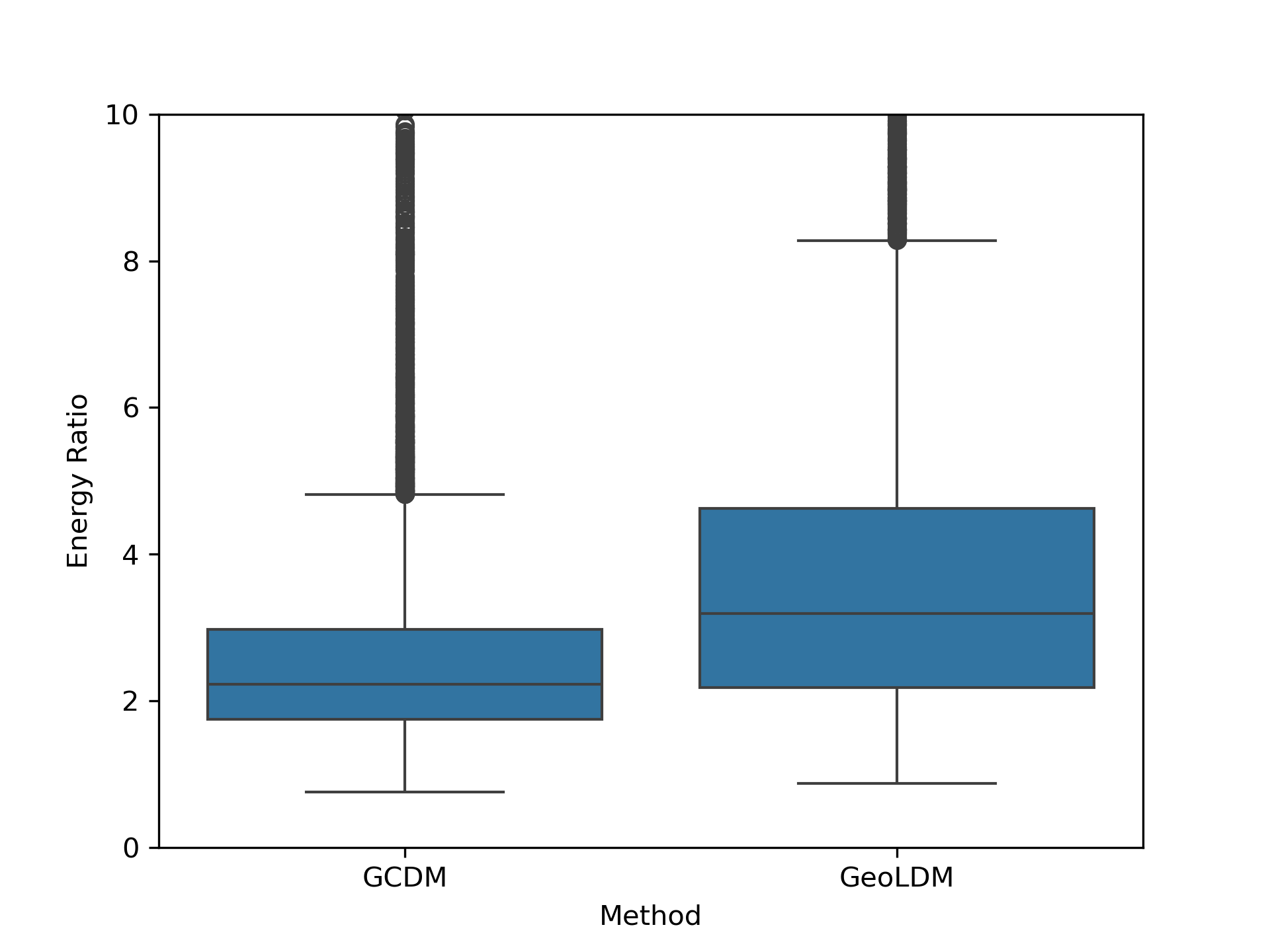}
\caption{A comparison of the energy ratios \citep{buttenschoen2024posebusters} of 10,000 large 3D molecules generated by \textsc{GCDM} and GeoLDM, a baseline state-of-the-art method. Employing Student's t-distribution 95\% confidence intervals, \textsc{GCDM} achieves a mean energy ratio of 2.98 $\pm$ 0.13, whereas GeoLDM yields a mean energy ratio of 4.19 $\pm$ 0.09.}
\label{figure:gcdm_geoldm_geom_energy_ratios}
\end{figure}

Figure \ref{fig:unconditional_geom_samples} illustrates PoseBusters-valid examples of large molecules generated by \textsc{GCDM} at the scale of GEOM-Drugs, with the following corresponding SMILES strings from left to right:
\textbf{(\underline{a})} \textrm{CC(C)=N[N]C(=O)O[C]([CH]C(=O)NCCCCc1cccnc1)Cc1ccc2c(c1)OCO2},
\textbf{(\underline{b})} \textrm{CN(N)Cc1cccnc1C(=O)NCCCc1ccc(F)cc1},
\textbf{(\underline{c})} \textrm{C=CCC(=O)c1cc(C(N)=O)c2ccccc2n1},
\textbf{(\underline{d})} \textrm{CC(=O)N/N=C/N=C/C=C\textbackslash N=C(/O)[C](O)CC(=O)N(O)Cc1ccc(F)c(F)c1},
\textbf{(\underline{e})} \textrm{COC(=O)/C(CN)=C(\textbackslash [CH]c1cc(C(C)=O)c(C)n1C)c1cc(Cl)ccc1O}, and
\textbf{(\underline{f})} \textrm{CC[C@@H](C)/N=C/[C](N[N+](=O)[O-])C(=O)c1ccc(C(=O)O)cc1}. As an example of the notion that \textsc{GCDM} produces low energy structures for a generated molecular graph, the free energies for Figures \ref{fig:unconditional_geom_samples} (a) and (f) were computed to be -3 kcal/mol and -2 kcal/mol, respectively, using \textsc{CREST} \citep{pracht2020automated} at the \textsc{GFN2-xTB} level of theory (which matches the corresponding free energy distribution mean for the GEOM-Drugs dataset (-2.5 kcal/mol) as illustrated in Figure 2 of \citep{axelrod2022geom} ). Lastly, to detect whether a method, in aggregate, generates molecules with unlikely 3D conformations, a generated molecule's energy ratio is defined as in \citet{buttenschoen2024posebusters} to be the ratio of the molecule's UFF-computed energy \citep{rappe1992uff} and the mean of 50 RDKit ETKDGv3-generated conformers \citep{riniker2015better} of the same molecular graph. Note that, as discussed by \citet{wills2023fragment}, generated molecules with an energy ratio greater than 7 are considered to have highly unlikely 3D conformations. Subsequently, Figure \ref{figure:gcdm_geoldm_geom_energy_ratios} reveals that the average energy ratio of \textsc{GCDM}'s large 3D molecules is notably lower and more tightly bounded compared to GeoLDM, the baseline SOTA method for this task, indicating that \textsc{GCDM} also generates more energetically-stable 3D molecule conformations compared to prior methods.

\subsection{Property-Guided 3D Molecule Optimization - QM9}
\label{section:optimization_qm9}

\begin{figure}[t]
\centering
\includegraphics[width=\textwidth]{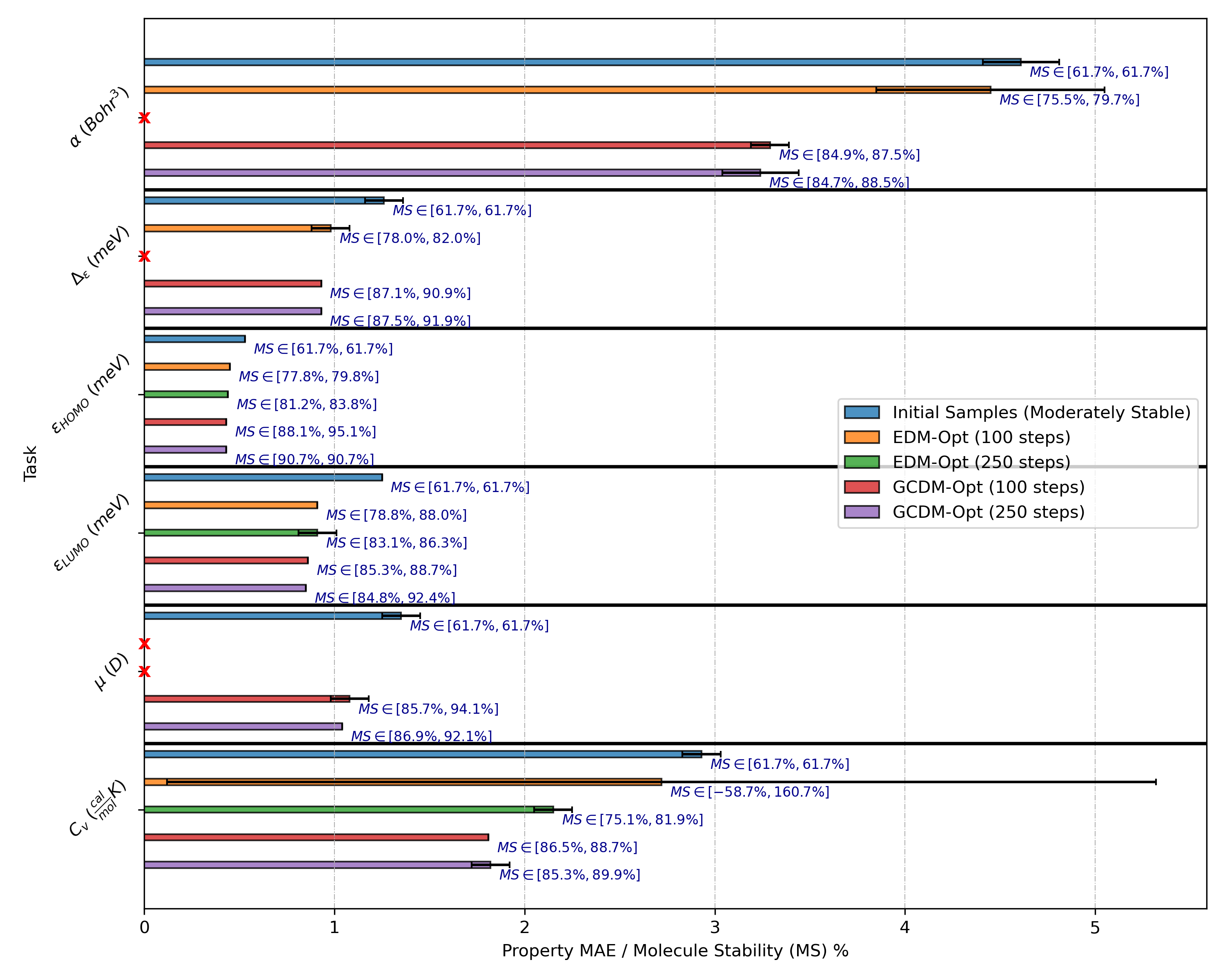}
\caption{Comparison of \textsc{GCDM} with baseline methods for property-guided 3D molecule optimization. The results are reported in terms of molecular stability ($MS$) and the MAE for molecular property prediction by an ensemble of three EGNN classifiers $\phi_{c}$ (each trained on the same QM9 subset using a distinct random seed) yielding corresponding Student's t-distribution 95\% confidence intervals, with results listed for EDM and \textsc{GCDM}-optimized samples as well as the molecule generation baseline ("Initial Samples"). Note that \textcolor{red}{x} denotes a missing bar representing outlier property MAEs greater than 50. Alternatively, tabular results are given in Table \ref{table:3dmg_optimization_qm9_results} of the appendix.}
\label{fig:3dmg_optimization_qm9_results}
\end{figure}

To evaluate whether molecular diffusion models can not only generate new 3D molecules but can also optimize existing small molecules using molecular property guidance, we adopt the QM9 dataset for the following experiment. First, we use an unconditional \textsc{GCDM} model to generate 1,000 3D molecules using 10 time steps of time-scaled reverse diffusion (to leave such molecules in an \textit{unoptimized} state), and then we provide these molecules to a separate property-conditional diffusion model for optimization of the molecules towards the conditional model's respective property. This conditional model accepts these 3D molecules as intermediate states for 100 and 250 time steps of property-guided optimization of the molecules' atom types and 3D coordinates. Lastly, we repurpose our experimental setup from Section \ref{section:conditional_qm9} to score these optimized molecules using an ensemble of external property classifier models to evaluate (1) how much the optimized molecules' predicted property values have been improved for the respective property (first metric) and (2) whether and how much the optimized molecules' stability (as defined in Section \ref{section:unconditional_qm9}) has been changed during optimization (second metric).

\textbf{Baselines.} Baseline methods for this experiment include EDM \citep{hoogeboom2022equivariant} and \textsc{GCDM}, where both methods use similar experimental setups for evaluation. Our baseline methods also include property-specificity and molecule stability measures of the initial (unconditional) 3D molecules to demonstrate how much molecular diffusion models can modify or improve these existing 3D molecules in terms of how property-specific and stable they are. As in Section \ref{section:conditional_qm9}, property specificity is measured in terms of the corresponding property classifier's MAE for a given molecule with a targeted property value, reporting the mean and Student's t-distribution 95\% confidence interval for each property MAE across an ensemble of three corresponding classifiers. Molecular stability (i.e., Mol Stable (\%)), here abbreviated at $MS$, is defined as in Section \ref{section:unconditional_qm9}.

\textbf{Results.} Figure \ref{fig:3dmg_optimization_qm9_results} showcases a practical finding: geometric diffusion models such as \textsc{GCDM} can effectively be repurposed as 3D molecule optimization methods with minimal modifications, improving both a molecule's stability and property specificity. This finding empirically supports the idea that molecular denoising diffusion models approximate the Boltzmann distribution with the score function they learn \citep{zaidi2022pre} and therefore may be applied in the optimization stage of the typical drug discovery pipeline \citep{deore2019stages} to experiment with a wider range of potential drug candidates (post-optimization) more quickly than previously possible. Simultaneously, the baseline EDM method fails to consistently optimize the stability and property specificity of existing 3D molecules, which suggests that geometric methods such as \textsc{GCDM} are theoretically and empirically better suited for such tasks. Notably, on average, with 100 time steps \textsc{GCDM} improves the stability of the initial molecules by \underline{over 25\%} and their specificity for each molecular property by \underline{over 27\%}, whereas for the properties it can optimize with 100 time steps, EDM improves the stability of the molecules by \underline{13\%} and their property specificity by \underline{15\%}. Lastly, it is worth noting that increasing the number of optimization time steps from 100 to 250 steps inconsistently leads to further improvements to molecules' stability and property specificity, indicating that the optimization trajectory likely reaches a local minimum around 100 time steps and hence rationalizes reducing the required compute time for optimizing 1,000 molecules e.g., from 15 minutes (for 250 steps) to 5 minutes (for 100 steps).

\subsection{Protein-Conditional 3D Molecule Generation}
\label{section:protein_conditional_experiment}

To investigate whether geometry-complete methods can enhance the ability of molecular diffusion models to generate 3D models within a given protein pocket (i.e., to perform structure-based drug design (SBDD)), in this experiment, we adopt the standard Binding MOAD (BM) \citep{hu2005binding} and CrossDocked (CD) \citep{francoeur2020three} datasets for training and evaluation of \textsc{GCDM-SBDD}, our geometry-complete, diffusion generative model based on \textsc{GCPNet++} that extends the diffusion framework of \citet{schneuing2022structure} for protein pocket-aware molecule generation. The Binding MOAD dataset consists of 100,000 high-quality protein-ligand complexes for training and 130 proteins for testing, with a 30\% sequence identity threshold being used to define this cross-validation split. Similarly, the CrossDocked dataset contains 40,484 high-quality protein-ligand complexes split between training (40,354) and test (100) partitions using proteins' enzyme commission numbers as described by \citet{schneuing2022structure}.

\textbf{Baselines.} Baseline methods for this experiment include DiffSBDD-cond \citep{schneuing2022structure} and DiffSBDD-joint \citep{schneuing2022structure}. We compare these methods to our proposed geometry-complete protein-aware diffusion model, \textsc{GCDM-SBDD}, using metrics that assess the properties, and thereby the quality, of each method's generated molecules. These molecule-averaged metrics include a method's average Vina score (computed using QuickVina 2.1) \citep{alhossary2015fast} as a physics-based estimate of a ligand's estimated binding affinity with a target protein, measured in units of kcal/mol (lower is better); average drug likeliness QED \citep{bickerton2012quantifying} (computed using RDKit 2022.03.2); average synthesizability \citep{ertl2009estimation} (computed using the procedure introduced by \citep{peng2022pocket2mol}) as an increasing measure of the ease of synthesizing a given molecule (higher is better); on average how many rules of Lipinski's rule of five are satisfied by a ligand \citep{lipinski2004lead} (computed compositionally using RDKit 2022.03.2); and average diversity in mean pairwise Tanimoto distances \citep{tanimoto1958elementary, bajusz2015tanimoto} (derived manually using fingerprints and Tanimoto similarities computed by RDKit 2022.03.2). Following established conventions for 3D molecule generation \citep{hoogeboom2022equivariant}, the size of each ligand to generate was determined using the ligand size distribution of the respective training dataset. Note that, in this context, "joint" and "cond" configurations represent generating a molecule for a protein target, respectively, with and without also modifying the coordinates of the binding pocket within the protein target. Also note that, similar to our experiments in Sections \ref{section:unconditional_qm9} - \ref{section:optimization_qm9}, the \textsc{GCDM-SBDD} model uses 9 \textsc{GCP} message-passing layers along with 256 (64) and 32 (16) invariant (equivariant) node and edge features, respectively.

\begin{table}
\centering
\resizebox{\textwidth}{!}{%
    \begin{tabular}{llcccccc}
        \toprule
        Dataset & Method & Vina (kcal/mol, $\downarrow$) & QED ($\uparrow$) & SA ($\uparrow$) & Lipinski ($\uparrow$) & Diversity ($\uparrow$) & PB-Valid (\%) ($\uparrow$) \\
        \midrule
        BM & DiffSBDD-cond (C$\alpha$) & $-5.784 \pm 0.03$ & $0.433 \pm 0.00$ & $0.616 \pm 0.00$ & $4.719 \pm 0.01$ & $0.848 \pm 0.00$ & $16.6 \pm 0.6\ /\ 1.7 \pm 0.2$ \\
        & DiffSBDD-joint (C$\alpha$) & $-5.882 \pm 0.05$ & $0.474 \pm 0.00$ & $0.631 \pm 0.00$ & $4.835 \pm 0.01$ & $0.852 \pm 0.00$ & $10.7 \pm 0.5\ /\ 0.7 \pm 0.1$ \\
        \rowcolor[gray]{0.8} & \textsc{GCDM-SBDD}-cond (C$\alpha$) (Ours) & $\textbf{-6.250} \pm 0.03$ & $\underline{0.465} \pm 0.00$ & $\underline{0.618} \pm 0.00$ & $4.661 \pm 0.01$ & $0.806 \pm 0.00$ & $\textbf{40.8} \pm 0.8\ /\ \textbf{6.8} \pm 0.4$ \\
        \rowcolor[gray]{0.8} & \textsc{GCDM-SBDD}-joint (C$\alpha$) (Ours) & $\underline{-6.159} \pm 0.06$ & $0.459 \pm 0.00$ & $0.584 \pm 0.00$ & $4.609 \pm 0.02$ & $0.794 \pm 0.00$ & $\underline{37.3} \pm 0.8\ /\ \underline{2.0} \pm 0.2$ \\
        & \textit{Reference} & $-8.328 \pm 0.04$ & $0.602 \pm 0.00$ & $0.336 \pm 0.00$ & $4.838 \pm 0.01$ & -- & -- \\
        \midrule
        CD & DiffSBDD-cond (C$\alpha$) & $-5.540 \pm 0.03$ & $0.449 \pm 0.00$ & $0.636 \pm 0.00$ & $4.735 \pm 0.01$ & $0.818 \pm 0.00$ & $40.7 \pm 1.0\ /\ 12.4 \pm 0.6$ \\
        & DiffSBDD-joint (C$\alpha$) & $-5.735 \pm 0.05$ & $0.420 \pm 0.00$ & $0.662 \pm 0.00$ & $4.859 \pm 0.01$ & $0.890 \pm 0.00$ & $34.1 \pm 0.9\ /\ 6.2 \pm 0.5$ \\
        \rowcolor[gray]{0.8} & \textsc{GCDM-SBDD}-cond (C$\alpha$) (Ours) & $\textbf{-5.955} \pm 0.04$ & $\underline{0.457} \pm 0.00$ & $\underline{0.640} \pm 0.00$ & $\underline{4.758} \pm 0.02$ & $0.795 \pm 0.00$ & $38.1 \pm 1.0\ /\ \textbf{15.7} \pm 0.7$ \\
        \rowcolor[gray]{0.8} & \textsc{GCDM-SBDD}-joint (C$\alpha$) (Ours) & $\underline{-5.870} \pm 0.03$ & $\textbf{0.458} \pm 0.00$ & $0.631 \pm 0.00$ & $4.701 \pm 0.02$ & $0.810 \pm 0.00$ & $\textbf{46.8} \pm 1.0\ /\ 6.5 \pm 0.5$ \\
        & \textit{Reference} & $-6.871 \pm 0.04$ & $0.476 \pm 0.00$ & $0.728 \pm 0.00$ & $4.340 \pm 0.00$ & -- & -- \\
        \bottomrule
    \end{tabular}%
}
\caption{Evaluation of generated molecules for target protein pockets from the Binding MOAD (BM) and CrossDocked (CD) test datasets. Our proposed method, \textsc{GCDM-SBDD}, achieves the best results for the metrics listed in \textbf{bold} and the second-best results for the metrics \underline{underlined}. For each metric, a method's mean and Student's t-distribution 95\% confidence error interval ($\pm$) is reported over 100 generated molecules for each test pocket. Additionally, the PoseBusters validity (PB-Valid) metric is defined as the percentage of generated molecules that pass all \textit{docking}-relevant structural and chemical sanity checks proposed by \cite{buttenschoen2024posebusters}, with the validity ratio to the left (right) of each $/$ denoting the percentage of valid molecules without (with) consideration of protein-ligand steric clashes.}
\label{table:3dmg_protein_conditional_results}
\end{table}

\begin{figure}[hbt!]
  \centering
  
  \subfloat[\textsc{2FKY}: -8.7 Vina]{\fbox{\includegraphics[height=1.7cm]{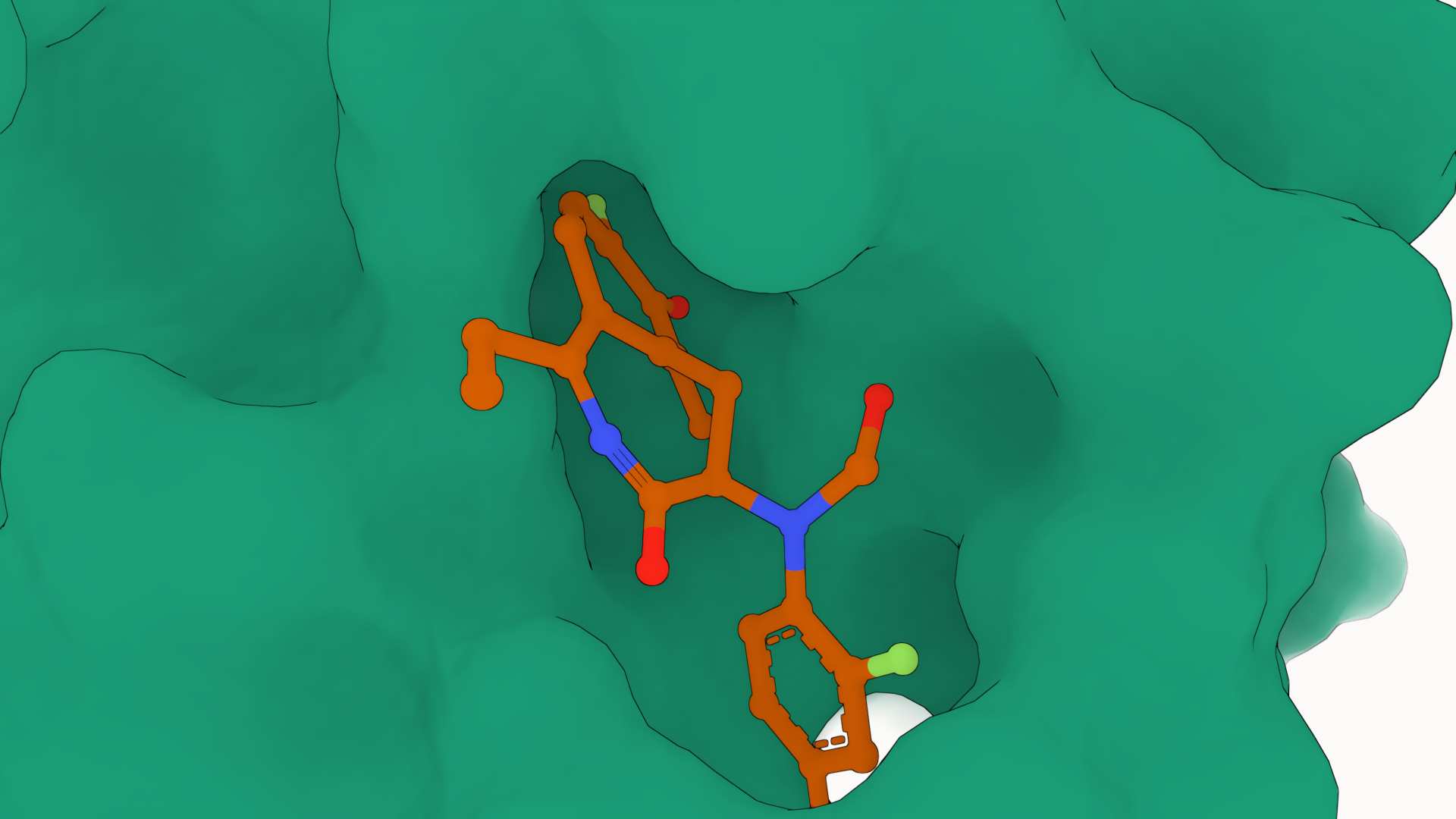}}}
  \subfloat[\textsc{2FKY}: -9.7 Vina]{\fbox{\includegraphics[height=1.7cm]{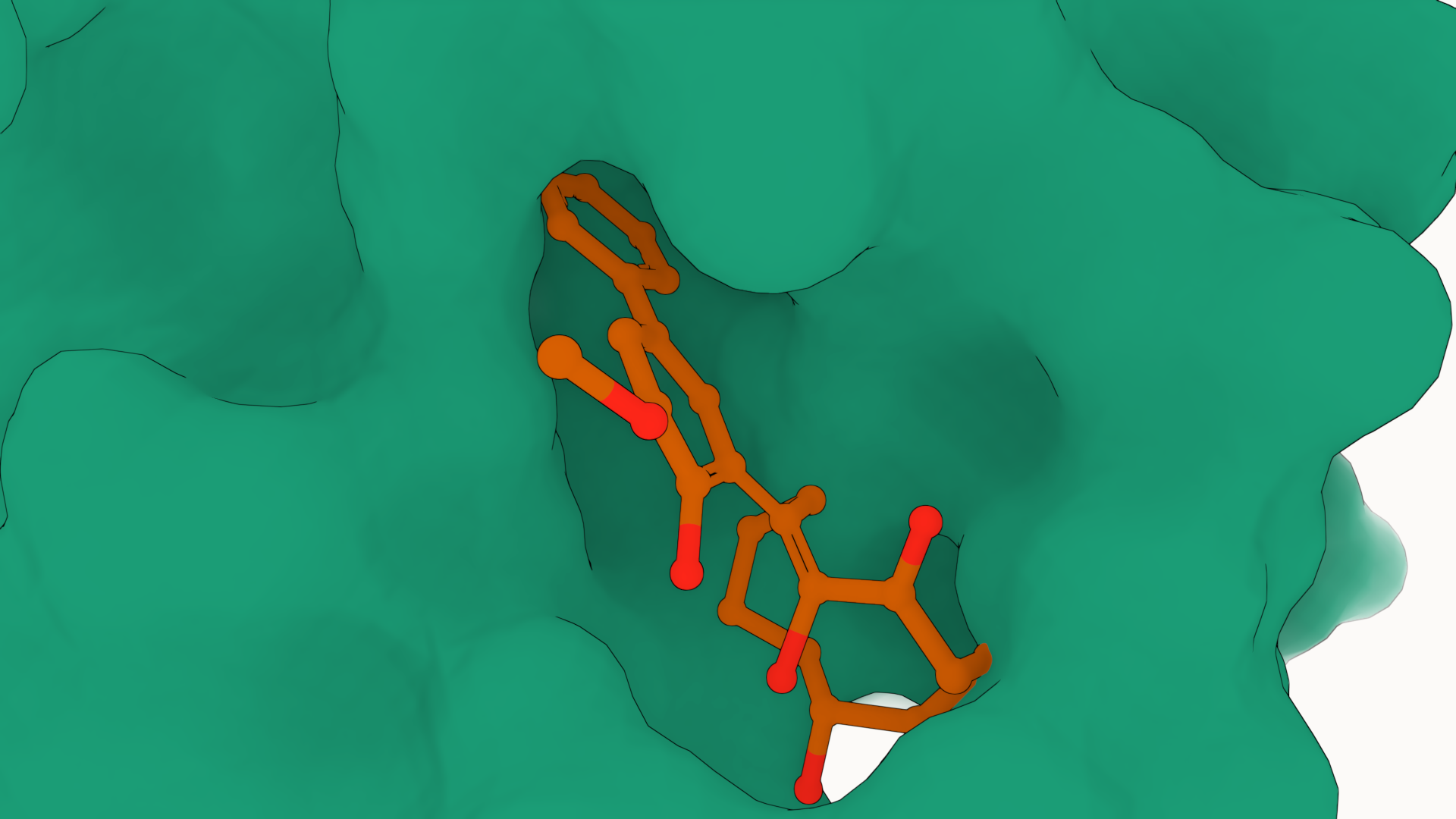}}}
  \subfloat[\textsc{4Z2G}: -8.2 Vina]{\fbox{\includegraphics[height=1.7cm]{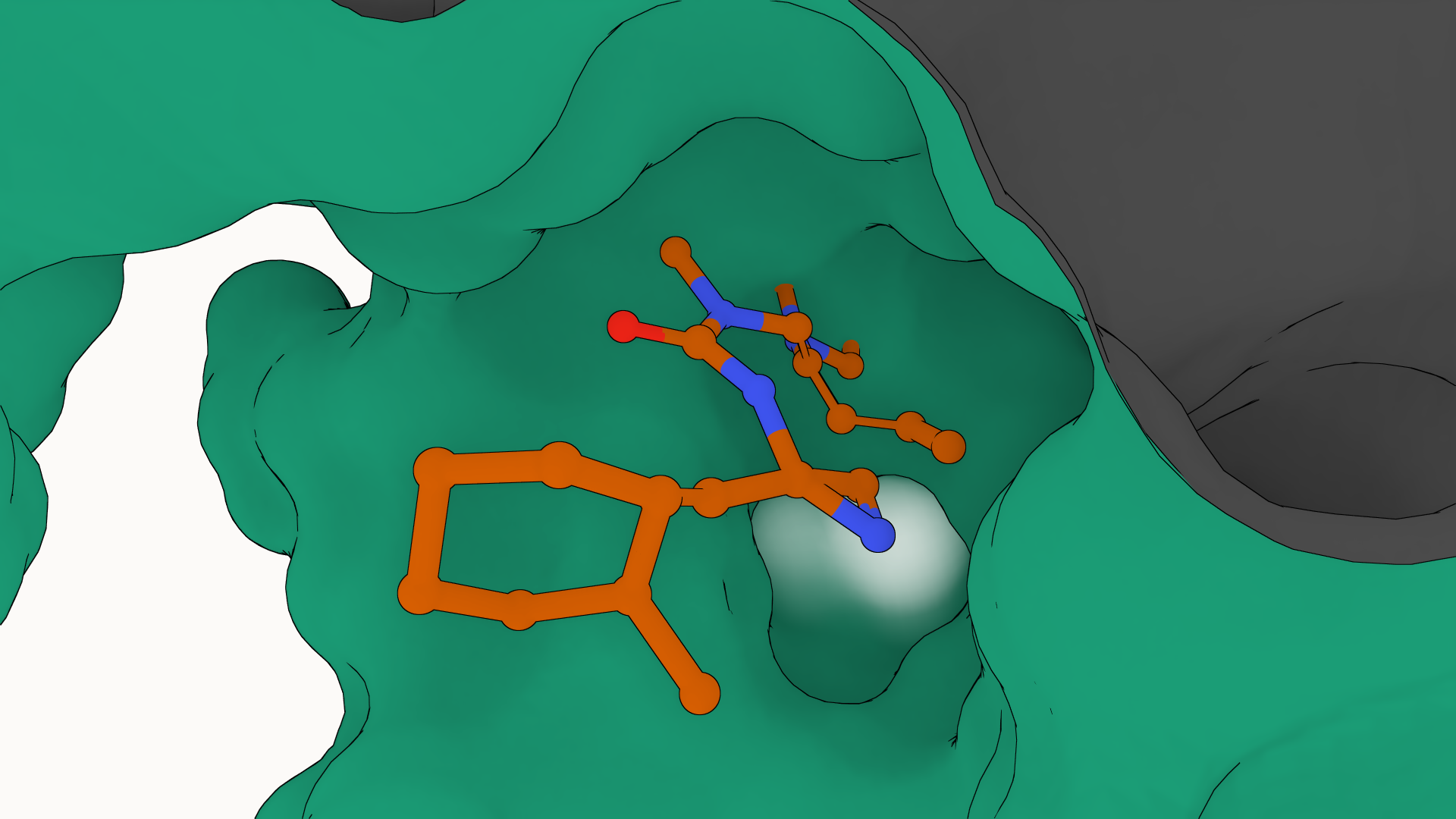}}}
  \subfloat[\textsc{4Z2G}: -8.7 Vina]{\fbox{\includegraphics[height=1.7cm]{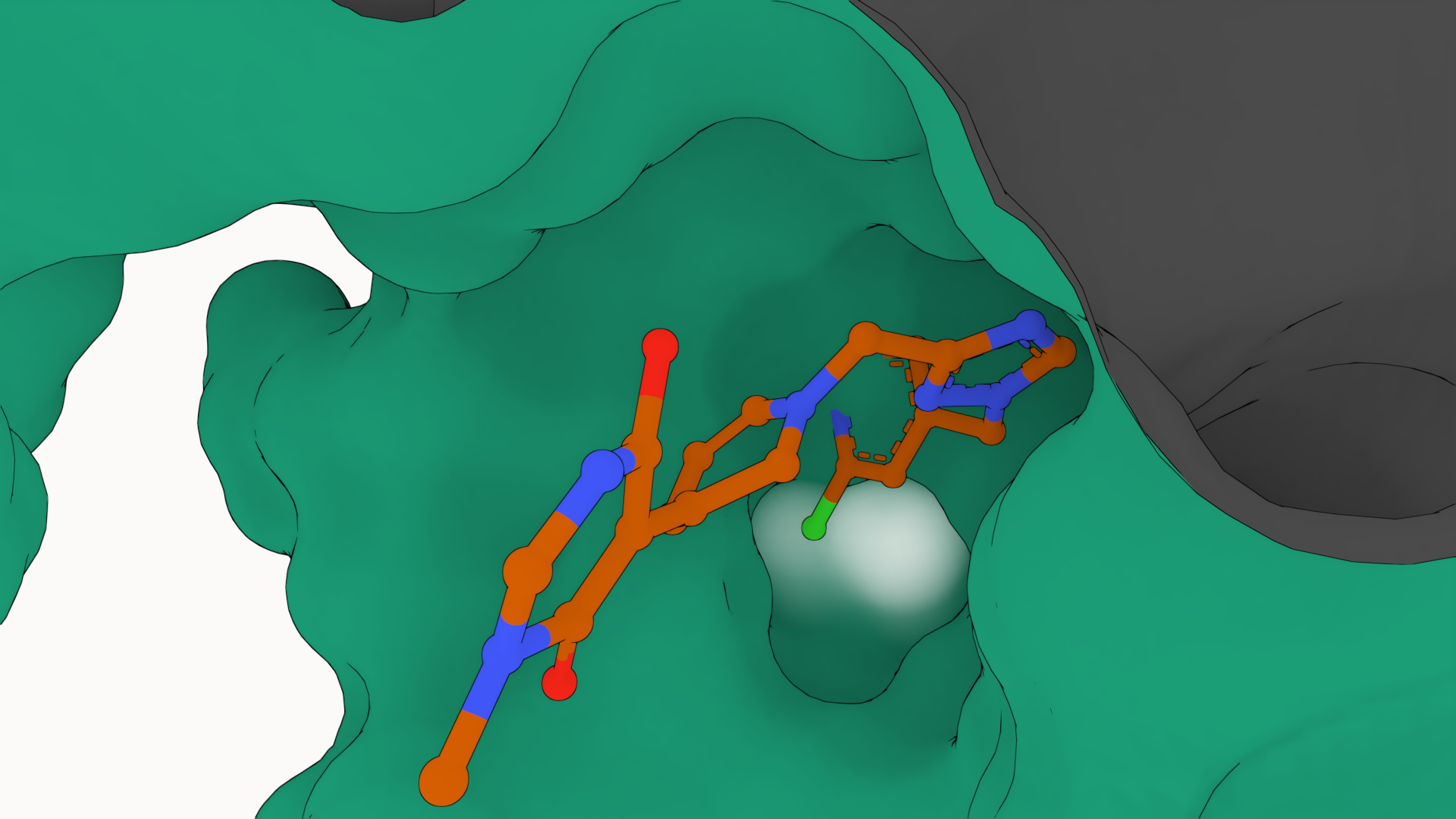}}}
  
  \caption{\textsc{GCDM-SBDD} molecules generated for BM (a-b) and CD (c-d) test proteins.}
  \label{fig:protein_conditional_samples}
\end{figure}

\textbf{Results.} Table \ref{table:3dmg_protein_conditional_results} shows that, across both of the standard SBDD datasets (i.e., Binding MOAD and CrossDocked), \textsc{GCDM-SBDD} generates more clash-free (PB-Valid) and lower energy (Vina) molecules compared to prior methods. Moreover, across all other metrics, \textsc{GCDM-SBDD} achieves comparable or better results in terms of drug-likeness measures (e.g., QED) and comparable results for all other molecule metrics \textit{without performing any hyperparameter tuning due to compute constraints}. These results suggest that \textsc{GCDM}, with \textsc{GCPNet++} as its denoising neural network, not only works well for de novo 3D molecule generation but also protein target-specific 3D molecule generation, notably expanding the number of real-world application areas of \textsc{GCDM}. Concretely, \textsc{GCDM-SBDD} improves upon DiffSBDD's average Vina energy scores by 8\% on average across both datasets while generating more than twice as many PB-valid "candidate" molecules for the more challenging Binding MOAD dataset.

As suggested by \cite{buttenschoen2024posebusters}, the gap between the PB-Valid ratios in Table \ref{table:3dmg_protein_conditional_results} without and with protein-ligand steric clashes considered for both \textsc{GCDM-SBDD} and DiffSBDD suggests that deep learning-based drug design methods for targeted protein pockets can likely benefit significantly from interaction-aware molecular dynamics relaxation following protein-conditional molecule generation, which may allow for many generated "candidate" molecules to have their PB validity "recovered" by such relaxation. Nonetheless, Figure \ref{fig:protein_conditional_samples} demonstrates that \textsc{GCDM} can consistently generate clash-free realistic and diverse 3D molecules with low Vina energies for unseen protein targets.

\section{Methods}
\label{section:methods}
\subsection{Problem Setting}
In this work, our goal is to generate new 3D molecules either unconditionally or conditioned on user-specified properties. We represent a molecular point cloud (e.g., 3D molecule) as a fully-connected 3D graph $\mathcal{G} = (\mathcal{V}, \mathcal{E})$ with $\mathcal{V}$ and $\mathcal{E}$ representing the graph's sets of nodes and edges, respectively, and $N = |\mathcal{V}|$ and $E = |\mathcal{E}|$ representing the numbers of nodes and edges in the graph, accordingly. In addition, $\mathbf{X} = (\mathbf{x}_{1}, \mathbf{x}_{2}, ..., \mathbf{x}_{N}) \in \mathbb{R}^{N \times 3}$ represents the respective Cartesian coordinates for each node (i.e., atom). Each node in $\mathcal{G}$ is described by scalar features $\mathbf{H} \in \mathbb{R}^{N \times h}$ and $m$ vector-valued features $\bm{\chi} \in \mathbb{R}^{N \times (m \times 3)}$. Likewise, each edge in $\mathcal{G}$ is described by scalar features $\mathbf{E} \in \mathbb{R}^{E \times e}$ and $x$ vector-valued features $\bm{\xi} \in \mathbb{R}^{E \times (x \times 3)}$. Then, let $\mathcal{M} = [\mathbf{X}, \mathbf{H}]$ represent the molecules (i.e., atom coordinates and atom types) our method is tasked with generating, where $[\cdot, \cdot]$ denotes the concatenation of two variables. Important to note is that the input features $\mathbf{H}$ and $\mathbf{E}$ are \textit{invariant} to 3D roto-translations, whereas the input vector features $\mathbf{X}$, $\bm{\chi}$ and $\bm{\xi}$ are \textit{equivariant} to 3D roto-translations. Lastly, in particular, we design a denoising neural network $\bm{\Phi}$ to be equivariant to 3D roto-translations (i.e., \textit{SE}(3)-equivariant) by defining it such that its internal operations and outputs match corresponding 3D roto-translations acting upon its inputs.

\subsection{Overview of \textsc{GCDM}}
We will now introduce \textsc{GCDM}, a new Geometry-Complete \textrm{SE(3)}-Equivariant Diffusion Model. \textsc{GCDM} defines a joint noising process on equivariant atom coordinates $\mathbf{x}$ and invariant atom types $\mathbf{h}$ to produce a noisy representation $\mathbf{z} = [\mathbf{z}^{(\mathbf{x})}, \mathbf{z}^{(\mathbf{h})}]$ and then learns a generative \textit{denoising} process using the newly-proposed \textsc{GCPNet++} model (see Section \ref{section:gcpnet++} of the appendix), which desirably contains two distinct feature channels for scalar and vector features, respectively, and supports geometry-complete and chirality-aware message-passing \citep{morehead2024geometry}.

As an extension of the DDPM framework \citep{ho2020denoising} outlined in Appendix \ref{section:appendix_diffusion_models}, \textsc{GCDM} is designed to generate molecules in 3D while maintaining \textrm{SE(3)} equivariance, in contrast to previous methods that generate molecules solely in 1D \citep{segler2018generating}, 2D \citep{pmlr-v80-jin18a}, or 3D modalities without considering chirality \citep{hoogeboom2022equivariant, xu2022geodiff}. \textsc{GCDM} generates molecules by directly placing atoms in continuous 3D space and assigning them discrete types, which is accomplished by modeling forward and reverse diffusion processes, respectively:
\begin{align*}
\underbrace{q(\mathbf{z}_{1:T} | \mathbf{z}_{0})}_{\text{Forward}} = \prod_{t = 1}^{T} q(\mathbf{z}_{t} | \mathbf{z}_{t - 1}) \qquad \underbrace{p_{\bm{\Phi}}(\mathbf{z}_{0:T - 1} | \mathbf{z}_{T})}_{\text{Reverse}} = \prod_{t = 1}^{T} p_{\bm{\Phi}}(\mathbf{z}_{t - 1} | \mathbf{z}_{t})
\end{align*}
Overall, these processes describe a latent variable model $p_{\bm{\Phi}}(\mathbf{z}_{0}) = \int p_{\bm{\Phi}}(\mathbf{z}_{0:T})d\mathbf{z}_{1:T}$ given a sequence of latent variables $\mathbf{z}_{0}, \mathbf{z}_{1}, \dots, \mathbf{z}_{T}$ matching the dimensionality of the data $\mathcal{M} \sim p(\mathbf{z}_{0})$. As illustrated in Figure \ref{figure:gcdm}, the forward process (directed from right to left) iteratively adds noise to an input, and the learned reverse process (directed from left to right) iteratively denoises a noisy input to generate new examples from the original data distribution. We will now proceed to formulate \textsc{GCDM}'s joint diffusion process and its remaining practical details.

\subsection{Joint Molecular Diffusion}
\label{section:gcdm_diffusion_process}
Recall that our model's molecular graph inputs, $\mathcal{G}$, associate with each node a 3D position $\mathbf{x}_{i} \in \mathbb{R}^{3}$ and a feature vector $\mathbf{h}_{i} \in \mathbb{R}^{h}$. By way of adding random noise to these model inputs at each time step $t$ via a fixed, Markov chain variance schedule $\sigma_{1}^{2}, \sigma_{2}^{2}, \dots, \sigma_{T}^{2}$, we can define a joint molecular diffusion process for equivariant atom coordinates $\mathbf{x}$ and invariant atom types $\mathbf{h}$ as the product of two distributions \citep{hoogeboom2022equivariant}:
\begin{equation}
    \label{equation:latent_variables_concat}
    q(\mathbf{z}_{t} | \mathbf{z}_{t - 1}) = \mathcal{N}_{xh}(\mathbf{z}_{t} | \alpha_{t}\mathbf{z}_{t - 1}, \sigma_{t}^{2} \mathbf{I}).
\end{equation}
where $\mathcal{N}_{xh}$ serves as concise notation to denote the product of two normal distributions; the first distribution, $\mathcal{N}_{x}$, represents the noised node coordinates; the second distribution, $\mathcal{N}_{h}$, represents the noised node features; and $\alpha_{t} = \sqrt{1 - \sigma_{t}^{2}}$ following the variance preserving process of \citet{ho2020denoising}.
With $\alpha_{t | s} = \alpha_{t} / \alpha_{s}$ and $\sigma_{t | s}^{2} = \sigma_{t}^{2} - \alpha_{t | s} \sigma_{s}^{2}$ for any $t > s$, we can \underline{directly} obtain the noisy data distribution $q(\mathbf{z}_{t} | \mathbf{z}_{0})$ at any time step $t$:
\begin{equation}
    q(\mathbf{z}_{t} | \mathbf{z}_{0}) = \mathcal{N}_{xh}(\mathbf{z}_{t} | \alpha_{t | 0} \mathbf{z}_{0}, \sigma_{t | 0}^{2} \mathbf{I}).
\end{equation}

Bayes Theorem then tells us that if we then define $\bm{\mu}_{t \rightarrow s}(\mathbf{z}_{t}, \mathbf{z}_{0})$ and $\sigma_{t \rightarrow s}$ as
\begin{align*}
    \bm{\mu}_{t \rightarrow s}(\mathbf{z}_{t}, \mathbf{z}_{0}) =\frac{\alpha_{s} \sigma_{t | s}^{2}}{\sigma_{t}^{2}}\mathbf{z}_{0}\ + \frac{\alpha_{t | s} \sigma_{s}^{2}}{\sigma_{t}^{2}}\mathbf{z}_{t} \mbox{ and }\ \sigma_{t \rightarrow s} = \frac{\sigma_{t | s} \sigma_{s}}{\sigma_{t}},
\end{align*}
we have that the inverse of the noising process, the \textit{true denoising process}, is given by the posterior of the transitions conditioned on $\mathcal{M} \sim \mathbf{z}_{0}$, a process that is also Gaussian \citep{hoogeboom2022equivariant}:
\begin{equation}
    q(\mathbf{z}_{s} | \mathbf{z}_{t}, \mathbf{z}_{0}) = \mathcal{N}(\mathbf{z}_{s} | \bm{\mu}_{t \rightarrow s}(\mathbf{z}_{t}, \mathbf{z}_{0}), \sigma_{t \rightarrow s}^{2} \mathbf{I}).
\end{equation}

\subsection{Parametrization of the Reverse Process}
\label{section:parametrization_of_the_reverse_process}
\textbf{Noise parametrization.} We now need to define the learned generative reverse process that \textit{denoises} pure noise into realistic examples from the original data distribution. Towards this end, we can directly use the noise posteriors $q(\mathbf{z}_{s} | \mathbf{z}_{t}, \mathbf{z}_{0})$ of Eq. \ref{equation:true_denoising_process} in the appendix after sampling $\mathbf{z}_{0} \sim (\mathcal{M} = [\mathbf{x}, \mathbf{h}])$. However, to do so, we must replace the input variables $\mathbf{x}$ and $\mathbf{h}$ with the approximations $\hat{\mathbf{x}}$ and $\hat{\mathbf{h}}$ predicted by the denoising neural network $\bm{\Phi}$:
\begin{equation}
    \label{equation:gcdm_denoising_equation}
    p_{\bm{\Phi}}(\mathbf{z}_{s} | \mathbf{z}_{t}) = \mathcal{N}_{xh}(\mathbf{z}_{s} | \bm{\mu}_{\bm{\Phi}_{t \rightarrow s}}(\mathbf{z}_{t}, \mathbf{\Tilde{z}}_{0}), \sigma_{t \rightarrow s}^{2} \mathbf{I}),
\end{equation}
where the values for $\mathbf{\Tilde{z}}_{0} = [\hat{\mathbf{x}}, \hat{\mathbf{h}}]$ depend on $\mathbf{z}_{t}$, $t$, and the denoising neural network $\bm{\Phi}$. \textsc{GCDM} then parametrizes $\bm{\mu}_{\bm{\Phi}_{t \rightarrow s}}(\mathbf{z}_{t}, \mathbf{\Tilde{z}}_{0})$ to predict the noise $\hat{\bm{\epsilon}} = [\hat{\bm{\epsilon}}^{(x)}, \hat{\bm{\epsilon}}^{(h)}]$, which represents the noise individually added to $\hat{\mathbf{x}}$ and $\hat{\mathbf{h}}$. We can then use the predicted $\hat{\bm{\epsilon}}$ to derive:
\begin{equation}
    \label{equation:noise_parametrization_outputs}
    \mathbf{\Tilde{z}}_{0} = [\hat{\mathbf{x}}, \hat{\mathbf{h}}] = \mathbf{z}_{t} / \alpha_{t} - \hat{\bm{\epsilon}}_{t} \cdot \sigma_{t} / \alpha_{t}.
\end{equation}
\textbf{Invariant likelihood.} Ideally, we desire for a 3D molecular diffusion model to assign the same likelihood to a generated molecule even after arbitrarily rotating or translating it in 3D space. To ensure the model achieves this desirable property for $p_{\bm{\Phi}}(\mathbf{z}_{0})$, we can leverage the insight that an invariant distribution composed of an equivariant transition function yields an invariant distribution \citep{satorras2021f, xu2022geodiff, hoogeboom2022equivariant}. Moreover, to address the translation invariance issue raised by \citet{satorras2021f} in the context of handling a distribution over 3D coordinates, we adopt the zero center of gravity trick proposed by \citet{xu2022geodiff} to define $\mathcal{N}_{x}$ as a normal distribution on the subspace defined by $\sum_{i}\mathbf{x}_{i} = \mathbf{0}$. In contrast, to handle node features $\mathbf{h}_{i}$ that are invariant to roto-translations, we can instead use a conventional normal distribution $\mathcal{N}$. As such, if we parametrize the transition function $p_{\bm{\Phi}}$ using an \textrm{SE(3)}-equivariant neural network after using the zero center of gravity trick of \citet{xu2022geodiff}, the model will have achieved the desired likelihood invariance property.

\subsection{Geometry-Complete Denoising Network}
Crucially, to satisfy the desired likelihood invariance property described in Section \ref{section:parametrization_of_the_reverse_process} while optimizing for model expressivity and runtime, \textsc{GCDM} parametrizes the denoising neural network $\bm{\Phi}$ using \textsc{GCPNet++}, an enhanced version of the \textrm{SE(3)}-equivariant \textsc{GCPNet} algorithm \citep{morehead2024geometry} , that we propose in Section \ref{section:gcpnet++} of the appendix. Notably, \textsc{GCPNet++} learns both scalar (invariant) and vector (equivariant) node and edge features through a chirality-sensitive graph message passing procedure, which enables \textsc{GCDM} to denoise its noisy molecular graph inputs using not only noisy scalar features but also noisy \textit{vector} features that are derived \underline{directly} from the noisy node coordinates $\mathbf{z}^{(\mathbf{x})}$ (i.e., $\psi(\mathbf{z}^{(\mathbf{x})})$). We empirically find that incorporating such noisy vectors considerably increases \textsc{GCDM}'s representation capacity for 3D graph denoising.

\subsection{Optimization Objective} Following previous works on diffusion models \citep{ho2020denoising, hoogeboom2022equivariant, wu2022diffusion}, the noise parametrization chosen for \textsc{GCDM} yields the following model training objective:
\begin{equation}
    \mathcal{L}_{t} = \mathbb{E}_{\epsilon_{t} \sim \mathcal{N}_{xh}(0, 1)}\left[ 
\frac{1}{2} w(t) \lVert \epsilon_{t} - \hat{\epsilon}_{t} \rVert^{2} \right],
\end{equation}
where $\hat{\epsilon}_{t}$ is the denoising network's noise prediction for atom types and coordinates as described above and where we empirically choose to set $w(t) = 1$ for the best possible generation results. Additionally, \textsc{GCDM} permits a negative log-likelihood computation using the same optimization terms as \citet{hoogeboom2022equivariant}, for which we refer interested readers to Appendices \ref{section:appendix_likelihood_terms}, \ref{section:appendix_diffusion_models_and_equivariance}, and \ref{section:appendix_gcdm_training_and_sampling} of the appendix.

\section{Discussion \& Conclusions}\label{sec12}

While previous methods for 3D molecule generation have possessed insufficient geometric and molecular priors for scaling well to a variety of molecular datasets, in this work, we introduced a geometry-complete diffusion model (\textsc{GCDM}) that establishes a clear performance advantage over previous methods, generating more realistic, stable, valid, unique, and property-specific 3D molecules, while enabling the generation of many large 3D molecules that are energetically stable as well as chemically and structurally valid. Moreover, \textsc{GCDM} does so without complex modeling techniques such as latent diffusion, which suggests that \textsc{GCDM}'s results could likely be further improved by expanding upon these techniques \citep{xu2023geometric}. Although \textsc{GCDM}'s results here are promising, since it (like previous methods) requires fully-connected graph attention as well as 1,000 time steps to generate a high-quality batch of 3D molecules, using it to generate several thousand large molecules can take a notable amount of time (e.g., 15 minutes to generate 250 new large molecules). As such, future research with \textsc{GCDM} could involve adding new time-efficient graph construction or sampling algorithms \citep{song2020denoising} or exploring the impact of higher-order (e.g., type-2 tensor) yet efficient geometric expressiveness \citep{liao2024equiformerv} on 3D generative models to accelerate sample generation and increase sample quality. Furthermore, integrating additional external tools for assessing the quality and rationality of generated molecules \citep{harris2023benchmarking} is a promising direction for future work.

\bmhead{Code Availability}
The source code for \textsc{GCDM} is available at \href{https://github.com/BioinfoMachineLearning/Bio-Diffusion}{https://github.com/BioinfoMachineLearning/Bio-Diffusion}, and the source code for structure-based drug design experiments with \textsc{GCDM} is separately available at \href{https://github.com/BioinfoMachineLearning/GCDM-SBDD}{https://github.com/BioinfoMachineLearning/GCDM-SBDD}.

\bmhead{Data Availability}
The data required to train new \textsc{GCDM} models or reproduce our results are available under a Creative Commons Attribution 4.0 International Public License at \href{https://zenodo.org/record/7881981}{https://zenodo.org/record/7881981}. Additionally, all pre-trained model checkpoints are available under a Creative Commons Attribution 4.0 International Public License at \href{https://zenodo.org/record/10995319}{https://zenodo.org/record/10995319}.

\bmhead{Supplementary information}
This article has an accompanying supplementary file containing the appendix for the article's main text.

\bmhead{Acknowledgments}
The authors would like to thank Chaitanya Joshi and Roland Oruche for helpful discussions and feedback on early versions of this manuscript. In addition, the authors acknowledge that this work is partially supported by three NSF grants (DBI2308699, DBI1759934, and IIS1763246), two NIH grants (R01GM093123 and R01GM146340), three DOE grants (DE-AR0001213, DE-SC0020400, and DE-SC0021303), and the computing allocation on the Summit compute cluster provided by the Oak Ridge Leadership Computing Facility under Contract DE-AC05- 00OR22725.

\bmhead{Author Contributions Statement}
AM and JC conceived the project. AM designed the experiments. AM performed the experiments and collected the data. AM analyzed the data. JC secured the funding for this project. AM and JC wrote the manuscript. AM and JC edited the manuscript.

\bmhead{Competing Interests Statement}
The authors declare no competing interests.

\backmatter

\bibliography{GCDM}

\newpage
\begin{appendices}

\section{Expanded Discussion of Denoising}
\label{section:methods_appendix}
\subsection{Geometry-Complete Denoising}
\label{section:geometry_complete_denoising}
In this section, we postulate that certain types of geometric neural networks serve as more effective 3D graph denoising functions for molecular DDPMs. We describe this notion as follows.\\

\begin{hypothesis}
    \label{hypothesis:geometry_completeness}
    (Geometry-Complete Denoising)\textbf{.} \\ \\
    \textit{Geometric neural networks that achieve geometry-completeness are more robust in denoising 3D molecular network inputs compared to models that are not geometry-complete, in that geometry-complete methods unambiguously define direction-robust local geometric reference frames.\\}
\end{hypothesis}

This hypothesis comes as an extension of the definition of geometry-completeness from \citet{pmlr-v162-du22e} and \citet{morehead2024geometry}:\\

\begin{definition}
    \label{definition:3}
    (Geometric Completeness)\textbf{.} \\ \\
    Given a pair of node positions $(x_{i}^{t}, x_{j}^{t})$ in a 3D graph $\mathcal{G}$,\\
    with vectors $a_{ij}^{t} \in \mathbb{R}^{1 \times 3}$, $b_{ij}^{t} \in \mathbb{R}^{1 \times 3}$, and $c_{ij}^{t} \in \mathbb{R}^{1 \times 3}$ derived from $(x_{i}^{t}, x_{j}^{t})$, \\
    a local geometric representation $\bm{\mathcal{F}}_{ij}^{t} = (a_{ij}^{t}, b_{ij}^{t}, c_{ij}^{t}) \in \mathbb{R}^{3 \times 3}$ is considered\\
    geometrically complete if $\bm{\mathcal{F}}_{ij}^{t}$ is non-degenerate, hence forming \\ a
    \textit{local orthonormal basis} located at the tangent space of $x_{i}^{t}$. \\
\end{definition}

An intuition for the implications of Hypothesis \ref{hypothesis:geometry_completeness} and Definition \ref{definition:3} on molecular diffusion models is that geometry-complete networks should be able to more effectively learn the gradients of data distributions \citep{ho2020denoising} in which a global force field is present, as is typically the case with 3D molecules \citep{pmlr-v162-du22e}. This is because, broadly speaking, geometry-complete methods encode local reference frames for each node (or edge) under which the directions of arbitrary global force vectors can be mapped. In addition to describing the theoretical benefits offered to geometry-complete denoising networks, we support this hypothesis through specific ablation studies in Sections \ref{section:unconditional_qm9} and \ref{section:unconditional_geom} where we ablate the geometric frame encodings from \textsc{GCDM} and find that such frames are particularly useful in improving \textsc{GCDM}'s ability to generate realistic 3D molecules.

\subsection{GCPNet++}
\label{section:gcpnet++}
Inspired by its recent success in modeling 3D molecular structures with geometry-complete message-passing, we parametrize $p_{\bm{\Phi}}$ using an enhanced version of Geometry-Complete Perceptron Networks (\textsc{GCPNets}) that were originally introduced by \citet{morehead2024geometry}. To summarize, \textsc{GCPNet} is a geometry-complete graph neural network that is equivariant to \textrm{SE(3)} transformations of its graph inputs and maps nicely to the context of Hypothesis \ref{hypothesis:geometry_completeness}.

In this setting, with ($h_{i} \in \mathbf{H}$, $\chi_{i} \in \bm{\chi}$, $e_{ij} \in \mathbf{E}, \xi_{ij} \in \bm{\xi}$), \textsc{GCPNet++}, our enhanced version of \textsc{GCPNet}, consists of a composition of Geometry-Complete Graph Convolution ($\mathbf{GCPConv}$) layers $(h_{i}^{l}, \chi_{i}^{l}), x_{i}^{l} = \mathbf{GCPConv}[(h_{i}^{l - 1}, \chi_{i}^{l - 1}), (e_{ij}^{l - 1}, \xi_{ij}^{l - 1}), x_{i}^{l - 1}, \mathcal{F}_{ij}]$ which are defined as:
\begin{equation}
    n_{i}^{l} = \bm{\phi}^{l}(n_{i}^{l - 1}, \mathcal{A}_{\forall j \in \mathcal{N}(i)} \bm{\Omega}_{\omega}^{l}(n_{i}^{l - 1}, n_{j}^{l - 1}, e_{ij}^{l - 1}, \xi_{ij}^{l - 1}, \mathcal{F}_{ij})),
\end{equation}
where $n_{i}^{l} = (h_{i}^{l}, \chi_{i}^{l})$; $\bm{\phi}^{l}$ is a trainable function; $l$ signifies the representation depth of the network; $\mathcal{A}$ is a permutation-invariant aggregation function; $\bm{\Omega}_{\omega}$ represents a message-passing function corresponding to the $\omega$-th $\mathbf{GCP}$ message-passing layer \citep{morehead2024geometry}; and node $i$'s geometry-complete local frames are $\mathcal{F}_{ij}^{t} = (a_{ij}^{t}, b_{ij}^{t}, c_{ij}^{t})$, with $a_{ij}^{t} = \frac{x_{i}^{t} - x_{j}^{t}}{ \lVert x_{i}^{t} - x_{j}^{t} \rVert }, b_{ij}^{t} = \frac{x_{i}^{t} \times x_{j}^{t}}{ \lVert x_{i}^{t} \times x_{j}^{t} \rVert },$ and $c_{ij}^{t} = a_{ij}^{t} \times b_{ij}^{t}$, respectively. Importantly, \textsc{GCPNet++} restructures the network flow of $\mathbf{GCPConv}$ \citep{morehead2024geometry} for each iteration of node feature updates to simplify and enhance information flow, concretely from the form of
\begin{equation}
    \hat{n}^{l} = n^{l - 1} + f(\Omega_{\omega, v_{i}}^{l} | v_{i} \in \mathcal{V})
\end{equation}
to
\begin{equation}
    \hat{n}^{l} = n^{l - 1} \cup f((g_{e^{\omega}, v_{i}}^{l}, \Omega_{e^{\omega}, v_{i}}^{l}, \Omega_{\xi^{\omega}, v_{i}}^{l}) | v_{i} \in \mathcal{V})
\end{equation}
and from
\begin{equation}
    n^{l} = \textbf{ResGCP}_{r}^{l}(\tilde{n}_{r - 1}^{l})
\end{equation}
to
\begin{equation}
    n^{l} = \textbf{GCP}_{r}^{l}(\tilde{n}_{r - 1}^{l}).
\end{equation}
Note that here $f$ represents a summation or a mean function that is invariant to node order permutations; $\cup$ denotes the concatenation operation; $g^{l}_{e^{\omega}, v_{i}}$ represents the binary-valued (i.e., [0, 1]) output of a scalar message attention (gating) function, expressed as
\begin{equation}
    g^{l}_{e^{\omega}} = \sigma_{inf}(\phi_{inf}^{l}(\Omega_{e^{\omega}}^{l}))
\end{equation}
with $\phi_{inf} : \mathbb{R}^{e} \rightarrow [0, 1]^{1}$ mapping from high-dimensional scalar edge feature space to a single dimension and $\sigma$ denoting a sigmoid activation function; $r$ is the node feature update module index; $\textbf{ResGCP}$ is a version of the $\textbf{GCP}$ module with added residual connections; and $\Omega_{\omega, v_{i}}^{l} = (\Omega_{e^{\omega}, v_{i}}^{l}, \Omega_{\xi^{\omega}, v_{i}}^{l})$ represents the scalar ($e$) and vector-valued ($\xi$) messages derived with respect to node $v_{i}$ using up to $\omega$ message-passing iterations within each \textsc{GCPNet++} layer.

We found these adaptations to provide state-of-the-art molecule generation results compared to the original node feature updating scheme, which we found yielded sub-optimal results in the context of generative modeling. This highlights the importance of customizing representation learning algorithms for the generative modeling task at hand, since reasonable performance may not always be achievable with them without careful adaptations. It is worth noting that, since \textsc{GCPNet++} performs message-passing directly on 3D vector features, \textsc{GCDM} is thereby the \textbf{first} diffusion generative model that is in principle capable of generating 3D molecules with specific \textit{vector}-valued properties. We leave a full exploration of this idea for future work.

\subsection{Properties of GCDM}
\label{section:properties_of_gcdm}
If one desires to update the coordinate representations of each node in $\mathcal{G}$, as we do in the context of 3D molecule generation, the $\mathbf{GCPConv}$ module of \textsc{GCPNet++} provides a simple, \textrm{SE(3)}-equivariant method to do so using a dedicated $\mathbf{GCP}$ module as follows:
\begin{align}
    (h_{p_{i}}^{l}, \chi_{p_{i}}^{l}) &= \mathbf{GCP}_{p}^{l}(n_{i}^{l}, \mathcal{F}_{ij}) \\
    x_{i}^{l} &= x_{i}^{l - 1} + \chi_{p_{i}}^{l}, \mbox{where } \chi_{p_{i}}^{l} \in \mathbb{R}^{1 \times 3},
\end{align}
where $\mathbf{GCP}_{\cdot}^{l}(\cdot, \mathcal{F}_{ij})$ is defined to provide chirality-aware rotation and translation-invariant updates to $h_{i}$ and rotation-equivariant updates to $\chi_{i}$ following centralization of the input point cloud's coordinates $\mathbf{X}$ \citep{pmlr-v162-du22e}. The effect of using positional feature updates $\chi_{p_{i}}$ to update $x_{i}$ is, after decentralizing $\mathbf{X}$ following the final $\mathbf{GCPConv}$ layer, that updates to $x_{i}$ then become \textrm{SE(3)}-equivariant. As such, all transformations described above satisfy the required equivariance constraints. Therefore, in integrating \textsc{GCPNet++} as its 3D graph denoiser, \textsc{GCDM} achieves \textrm{SE(3)} equivariance, geometry-completeness, and likelihood invariance altogether. Important to note is that \textsc{GCDM} subsequently performs message-passing with vector features to denoise its geometric inputs, whereas previous methods denoise their inputs \textbf{solely} using geometrically-insufficient scalar message-passing \citep{joshi2023expressive} as we demonstrate through our experiments in Section \ref{section:results}.

\section{Expanded Discussion of Diffusion}

\subsection{Diffusion Models}
\label{section:appendix_diffusion_models}

Key to understanding the contributions in this work are denoising diffusion probabilistic models (DDPMs). As alluded to previously, once trained, DDPMs can generate new data of arbitrary shapes, sizes, formats, and geometries by learning to reverse a noising process acting on each model input. More precisely, for a given data point $\mathbf{x}$, a diffusion process adds noise to $\mathbf{x}$ for time step $t = 0, 1, ..., T$ to yield $\mathbf{z}_{t}$, a noisy representation of the input $\mathbf{x}$ at time step $t$. Such a process is defined by a multivariate Gaussian distribution:
\begin{equation}
    \label{equation:standard_denoising_distribution}
    q(\mathbf{z}_{t} | x) = \mathcal{N}(\mathbf{z}_{t} | \alpha_{t} \mathbf{x}_{t}, \sigma_{t}^{2} \mathbf{I}),
\end{equation}
where $\alpha_{t} \in \mathbb{R}^{+}$ regulates how much feature signal is retained and $\sigma_{t}^{2}$ modulates how much feature noise is added to input $\mathbf{x}$. Note that we typically model $\alpha$ as a function defined with smooth transitions from $\alpha_{0} = 1$ to $\alpha_{T} = 0$, where a special case of such a noising process, the variance preserving process \citep{sohl2015deep, ho2020denoising}, is defined by $\alpha_{t} = \sqrt{1 - \sigma_{t}^{2}}$. To simplify notation, in this work, we define the feature signal-to-noise ratio as SNR($t$)$\ = \alpha_{t}^{2} / \sigma_{t}^{2}$. Also interesting to note is that this diffusion process is Markovian in nature, indicating that we have transition distributions as follows:
\begin{equation}
    q(\mathbf{z}_{t} | \mathbf{z}_{s}) = \mathcal{N}(\mathbf{z}_{t} | \alpha_{t | s} \mathbf{z}_{s}, \sigma_{t | s}^{2} \mathbf{I}),
\end{equation}
for all $t > s$ with $\alpha_{t | s} = \alpha_{t} / \alpha_{s}$ and $\sigma_{t | s}^{2} = \sigma_{t}^{2} - \alpha_{t | s}^{2} \sigma_{s}^{2}$.
In total, then, we can write the noising process as:
\begin{equation}
    q(\mathbf{z}_{0}, \mathbf{z}_{1}, ..., \mathbf{z}_{T} | \mathbf{x}) = q(\mathbf{z}_{0} | x) \prod_{t = 1}^{T} q(\mathbf{z}_{t} | \mathbf{z}_{t - 1}).
\end{equation}
If we then define $\bm{\mu}_{t \rightarrow s}(\mathbf{x}, \mathbf{z}_{t})$ and $\sigma_{t \rightarrow s}$ as
\begin{align*}
    \bm{\mu}_{t \rightarrow s}(\mathbf{x}, \mathbf{z}_{t}) = \frac{\alpha_{t | s} \sigma_{s}^{2}}{\sigma_{t}^{2}}\mathbf{z}_{t} + \frac{\alpha_{s} \sigma_{t | s}^{2}}{\sigma_{t}^{2}}\mathbf{x}\ \mbox{ and }\ \sigma_{t \rightarrow s} = \frac{\sigma_{t | s} \sigma_{s}}{\sigma_{t}},
\end{align*}
we have that the inverse of the noising process, the \textit{true denoising process}, is given by the posterior of the transitions conditioned on $\mathbf{x}$, a process that is also Gaussian:
\begin{equation}
    \label{equation:true_denoising_process}
    q(\mathbf{z}_{s} | \mathbf{x}, \mathbf{z}_{t}) = \mathcal{N}(\mathbf{z}_{s} | \bm{\mu}_{t \rightarrow s}(\mathbf{x}, \mathbf{z}_{t}), \sigma_{t \rightarrow s} \mathbf{I}).
\end{equation}

\textbf{The Generative Denoising Process.} In diffusion models, we define the generative process according to the \textit{true denoising process}. However, for such a denoising process, we do not know the value of $\mathbf{x}$ \textit{a priori}, so we typically approximate it as $\hat{\mathbf{x}} = \phi(\mathbf{z}_{t}, t)$ using a neural network $\phi$. Doing so then lets us express the generative transition distribution $p(\mathbf{z}_{s} | \mathbf{z}_{t})$ as $q(\mathbf{z}_{s} | \hat{\mathbf{x}}(\mathbf{z}_{t}, t), \mathbf{z}_{t})$. As a practical alternative to Eq. \ref{equation:true_denoising_process}, we can represent this expression using the approximation for $\hat{\mathbf{x}}$:
\begin{equation}
    p(\mathbf{z}_{s} | \mathbf{z}_{t}) = \mathcal{N}(\mathbf{z}_{s} | \bm{\mu}_{t \rightarrow s}(\hat{\mathbf{x}}, \mathbf{z}_{t}), \sigma_{t \rightarrow s}^{2} \mathbf{I}).
\end{equation}
If we choose to define $s$ as $s = t - 1$, then we can derive the variational lower bound on the log-likelihood of $\mathbf{x}$ given the generative model as:
\begin{equation}
    \log p(\mathbf{x}) \ge \mathcal{L}_{0} + \mathcal{L}_{base} + \sum_{t = 1}^{T} \mathcal{L}_{t},
\end{equation}
where we note that $\mathcal{L}_{0} = \log p(\mathbf{x} | \mathbf{z}_{0})$ models the likelihood of the data given its noisy representation $\mathbf{z}_{0}$, $\mathcal{L}_{base} = -\mbox{KL}(q(\mathbf{z}_{T} | \mathbf{x}) | p(\mathbf{z}_{T}))$ models the difference between a standard normal distribution and the final latent variable $q(\mathbf{z}_{T} | \mathbf{x})$, and
\begin{align*}
    \mathcal{L}_{t} = -\mbox{KL}(q(\mathbf{z}_{s} | \mathbf{x}, \mathbf{z}_{t}) | p(\mathbf{z}_{s} | \mathbf{z}_{t})) \ \mbox{ for } t = 1, 2, ..., T.
\end{align*}
Note that, in this formation of diffusion models, the neural network $\phi$ directly predicts $\hat{\mathbf{x}}$. However, \citet{ho2020denoising} and others have found optimization of $\phi$ to be made much easier when instead predicting the Gaussian noise added to $\mathbf{x}$ to create $\hat{\mathbf{x}}$. An intuition for how this changes the neural network's learning dynamics is that, when predicting back the noise added to the model's input, the network is being trained to more directly differentiate which part of $\mathbf{z}_{t}$ corresponds to the input's feature signal (i.e., the underlying data point $\mathbf{x}$) and which part corresponds to added feature noise. In doing so, if we let $\mathbf{z}_{t} = \alpha_{t}\mathbf{x} + \sigma_{t}\bm{\epsilon}$, the neural network can then predict $\hat{\bm{\epsilon}} = \phi(\mathbf{z}_{t}, t)$ such that:
\begin{equation}
    \hat{\mathbf{x}} = (1 / \alpha_{t})\ \mathbf{z}_{t} - (\sigma_{t} / \alpha_{t})\ \hat{\bm{\epsilon}}.
\end{equation}
\citet{kingma2021variational} and others have since shown that, when parametrizing the denoising neural network in this way, the loss term $\mathcal{L}_{t}$ reduces to:
\begin{equation}
    \label{equation:noise_parametrization}
    \mathcal{L}_{t} = \mathbb{E}_{\bm{\epsilon} \sim \mathcal{N}(\mathbf{0}, \mathbf{I})} \left[\frac{1}{2}(1 - \mbox{SNR}(t - 1) / \mbox{SNR}(t)) \lVert\bm{\epsilon} - \hat{\bm{\epsilon}}\rVert^{2} \right]
\end{equation}
Note that, in practice, the loss term $\mathcal{L}_{base}$ should be close to zero when using a noising schedule defined such that $\alpha_{T} \approx 0$. Moreover, if and when $\alpha_{0} \approx 1$ \textit{and} $\mathbf{x}$ is a discrete value, we will find $\mathcal{L}_{0}$ to be close to zero as well.

\subsection{Zeroth Likelihood Terms for \textsc{GCDM} Optimization Objective}
\label{section:appendix_likelihood_terms}
For the zeroth likelihood terms corresponding to each type of input feature, we directly adopt the respective terms previously derived by \citet{hoogeboom2022equivariant}. Doing so enables a negative log-likelihood calculation for \textsc{GCDM}'s predictions. In particular, for integer node features, we adopt the zeroth likelihood term:
\begin{equation}
    \label{equation:node_integer_feature_likelihood}
    p(\mathbf{h} | \mathbf{z}_{0}^{(h)}) = \int_{\mathbf{h} - \frac{1}{2}}^{\mathbf{h} + \frac{1}{2}} \mathcal{N}(\mathbf{u} | \mathbf{z}_{0}^{(h)}, \sigma_{0}) \mbox{d}\mathbf{u},
\end{equation}
where we use the CDF of a standard normal distribution, $\Phi$, to compute Eq. \ref{equation:node_integer_feature_likelihood} as $\Phi((\mathbf{h} + \frac{1}{2} - \mathbf{z}_{0}^{(h)}) / \sigma_{0}) - \Phi((\mathbf{h} - \frac{1}{2} - \mathbf{z}_{0}^{(h)}) / \sigma_{0}) \approx 1$ for reasonable noise parameters $\alpha_{0}$ and $\sigma_{0}$ \citep{hoogeboom2022equivariant}. For categorical node features, we instead use the zeroth likelihood term:
\begin{equation}
    p(\mathbf{h} | \mathbf{z}_{0}^{(h)}) = C(\mathbf{h} | \mathbf{p}), \mathbf{p} \propto \int_{1 - \frac{1}{2}}^{1 + \frac{1}{2}} \mathcal{N}(\mathbf{u} | \mathbf{z}_{0}^{(h)}, \sigma_{0}) \mbox{d}\mathbf{u},
\end{equation}
where we normalize $\mathbf{p}$ to sum to one and where $C$ is a categorical distribution \citep{hoogeboom2022equivariant}. Lastly, for continuous node positions, we adopt the zeroth likelihood term:
\begin{equation}
    p(\mathbf{x} | \mathbf{z}_{0}^{(x)}) = \mathcal{N}\left(\mathbf{x} | \mathbf{z}_{0}^{(x)} / \alpha_{0} - \sigma_{0} / \alpha_{0} \hat{\bm{\epsilon}}_{0}, \sigma_{0}^{2} / \alpha_{0}^{2}\mathbf{I}\right)
\end{equation}
which gives rise to the log-likelihood component $\mathcal{L}_{0}^{(x)}$ as:
\begin{equation}
    \mathcal{L}_{0}^{(x)} = \mathbb{E}_{\bm{\epsilon}^{(x)} \sim \mathcal{N}_{x}(\mathbf{0}, \mathbf{I})}\left[\log Z^{-1} - \frac{1}{2} \lVert \bm{\epsilon}^{x} - \phi^{(x)}(\mathbf{z}_{0}, 0) \rVert^{2} \right],
\end{equation}
where $d = 3$ and the normalization constant $Z = (\sqrt{2\pi} \cdot \sigma_{0} / \alpha_{0})^{(N - 1) \cdot d}$ - in particular, its $(N - 1) \cdot d$ term - arises from the zero center of gravity trick mentioned in Section \ref{section:parametrization_of_the_reverse_process} of the main text \citep{hoogeboom2022equivariant}.

\subsection{Diffusion Models and Equivariant Distributions}
\label{section:appendix_diffusion_models_and_equivariance}
In the context of diffusion generative models of 3D data, one often desires for the marginal distribution $p(\mathbf{x})$ of their denoising neural network to be an invariant distribution. Towards this end, we observe that a conditional distribution $p(y | x)$ is equivariant to the action of 3D rotations by meeting the criterion:
\begin{equation}
    p(y | x) = p(\mathbf{R}y | \mathbf{R}x)\ \mbox{ for all orthogonal } \mathbf{R}.
\end{equation}
Moreover, a distribution is invariant to rotation transformations $\mathbf{R}$ when
\begin{equation}
    p(y) = p(\mathbf{R}y)\ \mbox{ for all orthogonal } \mathbf{R}.
\end{equation}
As \citet{kohler2020equivariant} and \citet{xu2022geodiff} have collectively demonstrated, we know that if $p(\mathbf{z}_{T})$ is invariant and the neural network we use to parametrize $p(\mathbf{z}_{t - 1} | \mathbf{z}_{t})$ is equivariant, we have, as desired, that the marginal distribution $p(\mathbf{x})$ of the denoising model is an invariant distribution.

\subsection{Training and Sampling Procedures for \textsc{GCDM}}
\label{section:appendix_gcdm_training_and_sampling}

\textbf{Equivariant Dynamics.} In this work, we use the previous definition of \textsc{GCPNet++} in Section \ref{section:gcpnet++} of the main text to learn an \textrm{SE(3)}-equivariant dynamics function $[\hat{\bm{\epsilon}}^{(x)}, \hat{\bm{\epsilon}}^{(h)}] = \phi(\mathbf{z}_{t}^{(x)}, \mathbf{z}_{t}^{(h)}, t)$ as:
\begin{equation}
    \hat{\bm{\epsilon}}_{t}^{(x)}, \hat{\bm{\epsilon}}_{t}^{(h)} = \textsc{GCPNet++}(\mathbf{z}_{t}^{(x)}, [\mathbf{z}_{t}^{(h)}, \psi(\mathbf{z}_{t}^{(x)}), t / T]) - [\mathbf{z}_{t}^{(x)}, \mathbf{0}],
\end{equation}
where we inform the denoising model of the current time step by concatenating $t / T$ as an additional node feature and where we subtract the coordinate representation outputs of \textsc{GCPNet++} from its coordinate representation inputs after subtracting from the coordinate representation outputs their collective center of gravity. Lastly yet importantly, as a geometric GNN, \textsc{GCPNet++} can embed geometric vector features in addition to scalar features. Subsequently, from the noisy coordinates representation $\mathbf{z}_{t}^{(x)}$ we derive noisy sequential (node) orientation unit vectors and pairwise (edge) displacement unit vectors $\psi(\mathbf{z}_{t}^{(x)})$, respectively, and embed these features using \textsc{GCPNet++}'s vector feature channels for nodes and edges accordingly. With the parametrization in Eq. \ref{equation:noise_parametrization_outputs} of the main text, \textsc{GCDM} subsequently achieves rotation equivariance on $\hat{\mathbf{x}}_{i}$, thereby achieving a 3D translation and rotation-invariant marginal distribution $p(\mathbf{x})$ as described in Appendix \ref{section:appendix_diffusion_models_and_equivariance}.

\textbf{Scaling Node Features.}
In line with \citet{hoogeboom2022equivariant}, to improve the log-likelihood of the model's generated samples, we find it useful to train and perform sampling with \textsc{GCDM} using scaled node feature inputs as $[\mathbf{x}, \frac{1}{4} \mathbf{h}^{(categorical)}, \frac{1}{10} \mathbf{h}^{(integer)}]$.

\textbf{Deriving The Number of Atoms.}
Finally, to determine the number of atoms with which \textsc{GCDM} will generate a 3D molecule, we first sample $N \sim p(N)$, where $p(N)$ denotes the categorical distribution of molecule sizes over \textsc{GCDM}'s training dataset. Then, we conclude by sampling $\mathbf{x}, \mathbf{h} \sim p(\mathbf{x}, \mathbf{h} | N)$.

\newpage
\section{Additional Details}
\label{section:appendix_additional_details}
\subsection{Broader Impacts}
In this work, we investigate the impact of geometric representation learning on generative models for 3D molecules. Such research can contribute to drug discovery efforts by accelerating the development of new medicinal or energy-related molecular compounds, and, as a consequence, can yield positive societal impacts \citep{walters2020assessing}. Nonetheless, in line with \citet{urbina2022dual}, we authors would argue that it will be critical for institutions, governments, and nations to reach a consensus on the strict regulatory practices that should govern the use of such molecule design methodologies in settings in which it is reasonably likely such methodologies could be used for nefarious purposes by scientific "bad actors".

\subsection{Training Details}
\label{section:appendix_training_details}
\textbf{Scalar Message Attention.} In our implementation of scalar message attention (SMA) within \textsc{GCDM}, $\mathbf{m}_{ij} = e_{ij} \mathbf{m}_{ij}$, where $\mathbf{m}_{ij}$ represents the scalar messages learned by \textsc{GCPNet++} during message-passing and $e_{ij}$ represents a 1 if an edge exists between nodes $i$ and $j$ (and a 0 otherwise) via $e_{ij} \approx \phi_{inf}(\mathbf{m}_{ij})$. Here, $\phi_{inf} : \mathbb{R}^{e} \rightarrow [0, 1]^{1}$ resembles a linear layer followed by a sigmoid function \citep{satorras2021n}.

\textbf{\textsc{GCDM} Hyperparameters.} All \textsc{GCDM} models train on QM9 for approximately 1,000 epochs using 9 $\mathbf{GCPConv}$ layers; SiLU activations \citep{elfwing2018sigmoid}; 256 and 64 scalar node and edge hidden features, respectively; and 32 and 16 vector-valued node and edge features, respectively. All \textsc{GCDM} models are also trained using the AdamW optimizer \citep{loshchilov2017decoupled} with a batch size of 64, a learning rate of $10^{-4}$, and a weight decay rate of $10^{-12}$.

\textbf{\textsc{GCDM} Runtime.} With a maximum batch size of 64, this 9-layer model configuration allows us to train \textsc{GCDM} models for unconditional (conditional) tasks on the QM9 dataset using approximately 10 (15) days of GPU training time with a single 24GB NVIDIA A10 GPU. For unconditional molecule generation on the much larger GEOM-Drugs dataset, a maximum batch size of 64 allows us to train 4-layer \textsc{GCDM} models using approximately 60 days of GPU training time with a single 48GB NVIDIA RTX A6000 GPU. As such, access to several GPUs with larger GPU memory limits (e.g., 80GBs) should allow one to concurrently train \textsc{GCDM} models in a fraction of the time via larger batch sizes or data-parallel training techniques \citep{falcon2019pytorch}.

\subsection{Compute Requirements}
Training \textsc{GCDM} models for tasks on the QM9 dataset by default requires a GPU with at least 24GB of GPU memory. Inference with such \textsc{GCDM} models for QM9 is much more flexible in terms of GPU memory requirements, as users can directly control how soon a molecule generation batch will complete according to the size of molecules being generated as well as one's selected batch size during sampling. Training \textsc{GCDM} models for unconditional molecule generation on the GEOM-Drugs dataset by default requires a GPU with at least 48GB of GPU memory. Similar to the \textsc{GCDM} models for QM9, inference with GEOM-Drugs models is flexible in terms of GPU memory requirements according to one's choice of sampling hyperparameters. Note that inference for both QM9 models and GEOM-Drugs models can likely be accelerated using techniques such as DDIM sampling \citep{song2020denoising}. However, we have not officially validated the quality of generated molecules using such sampling techniques, so we caution users to be aware of this potential risk of degrading molecule sample quality when using such sampling algorithms.

\subsection{Reproducibility}
On \href{https://github.com/BioinfoMachineLearning/Bio-Diffusion}{GitHub}, we thoroughly provide all source code, data, and instructions required to train new \textsc{GCDM} models or reproduce our results for each of the four protein-independent molecule generation tasks we study in this work. The source code, data, and instructions for our protein-conditional molecule generation experiments are also available on \href{https://github.com/BioinfoMachineLearning/GCDM-SBDD}{GitHub}. Our source code uses PyTorch \citep{paszke2019pytorch} and PyTorch Lightning \citep{falcon2019pytorch} to facilitate model training; PyTorch Geometric \citep{fey2019fast} to support sparse tensor operations on geometric graphs; and Hydra \citep{Yadan2019Hydra} to enable reproducible hyperparameter and experiment management.

\subsection{Additional Results}
\subsubsection{Property-Guided 3D Molecule Optimization - QM9}

\begin{table}[t]
\centering
\resizebox{\textwidth}{!}{%
    \begin{tabular}{lllllll}
        & & & & & & \\ \midrule
        Task & $\alpha \downarrow$ \textit{/} $MS \uparrow$ & $\Delta \epsilon \downarrow$ \textit{/} $MS \uparrow$ & $\epsilon_{HOMO} \downarrow$ \textit{/} $MS \uparrow$ & $\epsilon_{LUMO} \downarrow$ \textit{/} $MS \uparrow$ & $\mu \downarrow$ \textit{/} $MS \uparrow$ & $C_{v} \downarrow$ \textit{/} $MS \uparrow$ \\
        Units & $Bohr^{3}$ \textit{/} \% & $meV$ \textit{/} \% & $meV$ \textit{/} \% & $meV$ \textit{/} \% & $D$ \textit{/} \% & $\frac{cal}{mol} K$ \textit{/} \% \\ \midrule
        Initial Samples (Moderately Stable) & $4.61 \pm 0.2$ / 61.7 & $1.26 \pm 0.1$ / 61.7 & $0.53 \pm 0.0$ / 61.7 & $1.25 \pm 0.0$ / 61.7 & $1.35 \pm 0.1$ / 61.7 & $2.93 \pm 0.1$ / 61.7 \\ \midrule
        EDM-Opt (100 steps on initial samples) & $4.45 \pm 0.6$ / $77.6 \pm 2.1$ & $0.98 \pm 0.1$ / $80.0 \pm 2.0$ & $0.45 \pm 0.0$ / $78.8 \pm 1.0$ & $0.91 \pm 0.0$ / $83.4 \pm 4.6$ & $6e^{5} \pm 6e^{5}$ / $78.3 \pm 2.9$ & $2.72 \pm 2.6$ / $51.0 \pm 109.7$ \\
        EDM-Opt (250 steps on initial samples) & $1e^{2} \pm 5e^{2}$ / $80.1 \pm 2.1$ & $1e^{3} \pm 6e^{3}$ / $83.7 \pm 3.8$ & $0.44 \pm 0.0$ / $82.5 \pm 1.3$ & $0.91 \pm 0.1$ / $\underline{84.7} \pm 1.6$ & $2e^{5} \pm 8e^{5}$ / $\underline{81.0} \pm 5.8$ & $\underline{2.15} \pm 0.1$ / $\underline{78.5} \pm 3.4$ \\ \midrule
        \rowcolor[gray]{0.8} \textsc{GCDM}-Opt (100 steps on initial samples) & $\underline{3.29} \pm 0.1$ / $\underline{86.2} \pm 1.3$ & $\underline{0.93} \pm 0.0$ / $\underline{89.0} \pm 1.9$ & $\mathbf{0.43} \pm 0.0$ / $\mathbf{91.6} \pm 3.5$ & $\underline{0.86} \pm 0.0$ / $\underline{87.0} \pm 1.7$ & $\underline{1.08} \pm 0.1$ / $\mathbf{89.9} \pm 4.2$ & $\mathbf{1.81} \pm 0.0$ / $\underline{87.6} \pm 1.1$ \\
        \rowcolor[gray]{0.8} \textsc{GCDM}-Opt (250 steps on initial samples) & $\mathbf{3.24} \pm 0.2$ / $\mathbf{86.6} \pm 1.9$ & $\mathbf{0.93} \pm 0.0$ / $\mathbf{89.7} \pm 2.2$ & $\underline{0.43} \pm 0.0$ / $\underline{90.7} \pm 0.0$ & $\textbf{0.85} \pm 0.0$ / $\mathbf{88.6} \pm 3.8$ & $\mathbf{1.04} \pm 0.0$ / $\underline{89.5} \pm 2.6$ & $\underline{1.82} \pm 0.1$ / $\mathbf{87.6} \pm 2.3$ \\ \midrule
    \end{tabular}%
}
\caption{Comparison of \textsc{GCDM} with baseline methods for property-guided 3D molecule optimization. The results are reported in terms of molecular stability ($MS$) and the MAE for molecular property prediction by an ensemble of three EGNN classifiers $\phi_{c}$ (each trained on the same QM9 subset using a distinct random seed) yielding corresponding Student's t-distribution 95\% confidence intervals, with results listed for EDM and \textsc{GCDM}-optimized samples as well as the molecule generation baseline ("Initial Samples"). Note that certain experiments with an EDM optimizer yielded unsuccessful property optimization, where we denote such results as outlier property MAE values greater than 50. The top-1 (best) results for this task are in \textbf{bold}, and the second-best results are \underline{underlined}.}
\label{table:3dmg_optimization_qm9_results}
\end{table}

In Table \ref{table:3dmg_optimization_qm9_results}, for completeness, we list the numeric molecule optimization results comprising Figure \ref{fig:3dmg_optimization_qm9_results} in Section \ref{section:optimization_qm9}.




\end{appendices}


\end{document}